\documentclass[journal]{IEEEtran}

\ifCLASSINFOpdf
  \usepackage[pdftex]{graphicx}
  \DeclareGraphicsExtensions{.pdf,.jpeg,.png}
\else
  \usepackage[dvips]{graphicx}
  \DeclareGraphicsExtensions{.eps}
\fi
\usepackage{amsmath}
\usepackage{amssymb}
\usepackage{mathtools}

\usepackage[linesnumbered, ruled, vlined]{algorithm2e}
\usepackage {listings}
\newtheorem{definition}{Definition}

\usepackage{array}
\usepackage{caption}
\usepackage{subcaption}
\usepackage{float}

\usepackage{url}

\usepackage[all]{myglossary}

\usepackage{tabularx}
\usepackage{multirow}
\usepackage{longtable}
\usepackage{booktabs}
\usepackage[table,xcdraw]{xcolor}

\newcolumntype{L}[1]{>{\raggedright\let\newline\\\arraybackslash\hspace{0pt}}m{#1}}
\newcolumntype{C}[1]{>{\centering\let\newline\\\arraybackslash\hspace{0pt}}m{#1}}
\newcolumntype{R}[1]{>{\raggedleft\let\newline\\\arraybackslash\hspace{0pt}}m{#1}}

\begin{document}
%

\title{sKPNSGA-II: Knee point based MOEA with self-adaptive angle for Mission Planning Problems }

%
%
%

\author{Cristian~Ramirez-Atencia, 
        Sanaz~Mostaghim
        and~David~Camacho
\thanks{C. Ramirez-Atencia and S. Mostaghim are with faculty of Computer Science, Otto von Guericke University Magdeburg, Germany, e-mails: cristian.ramirez@ovgu.de, sanaz.mostaghim@ovgu.de.}
\thanks{D. Camacho is with the Information Systems Department, Technical University of Madrid, Spain, e-mail: david.camacho@etsisi.upm.es.}
}

\maketitle

\begin{abstract}
Real-world and complex problems have usually many objective functions that have to be optimized all at once. 
Over the last decades, \glspl{moea} are designed to solve this kind of problems. Nevertheless, some problems have many objectives which lead to a large number of non-dominated solutions obtained by the optimization algorithms. The large set of non-dominated solutions hinders the selection of the most appropriate solution by the decision maker. This paper presents a new algorithm that has been designed to obtain the most significant solutions from the \gls{pof}. This approach is based on the cone-domination applied to \gls{moea}, which can find the knee point solutions. In order to obtain the best cone angle, we propose a hypervolume-distribution metric, which is used to self-adapt the angle during the evolving process. This new algorithm has been applied to the real world application in \gls{uav} Mission Planning Problem. The experimental results show a significant improvement of the algorithm performance in terms of hypervolume, number of solutions, and also the required number of generations to converge.

\end{abstract}


%
\IEEEpeerreviewmaketitle
\glsresetall

\section{Introduction}
%
%
%
%
\IEEEPARstart{R}{eal}-world optimization problems often deal with multiple objectives that must be met simultaneously in the solutions. In most cases, objectives are conflicting, so improving one objective usually cannot be achieved unless other objective is worsened. Such problems are called \glspl{mop}, the solution of which is a set of solutions representing different performance trade-off between the objectives.

Most of the existing algorithms focus on the approximation of the \gls{pof} in terms of convergence and distribution, but always with a fixed population size, which in the end, for complex problems with many solutions, returns an amount of individuals equal or similar to this population. Nevertheless, when this approximation of the Pareto set comprises a large number of solutions, the process of decision making to select one appropriate solution becomes a difficult task for the \gls{dmaker}. Sometimes, the \gls{dmaker} provides a priori information about his/her preferences, which can be used in the optimization process \cite{Thiele2009,Goulart2016}. However, very often this information is not provided by the \gls{dmaker}, and it is necessary to consider other approaches to filter the number of solutions. In opposition to the common trend of returning a more or less fix amount of solutions, it should be taken into account the hardness of decision making for every additional solution, which in the end may not be optimal enough in comparison with other solutions. So, for complex real-world problems, it should be considered to provide a lower amount of solution maintaining as much as possible of the convergence and distribution of the \gls{pof}.

In the last years, finding the "knee points" \cite{Branke2004} have been used in several algorithms \cite{Setoguchi2015,Zhang2015} to deal with large \glspl{pof} in convex problems when the \gls{dmaker} does not provide preferences about the \gls{mop}. In this work
, a new \gls{moea} focused on the search of Knee Points is presented. This new algorithm changes the concept of domination to \textit{cone-domination}, where a larger portion (cone region) than a typical domination criteria is considered when the solution frontier is generated. In this paper, the main novelty with respect to a previous approach~\cite{Ramirez-Atencia2017} lies in the proposition of a new adaptive technique to find the right angle for the cone-domination which focus on reducing as much as possible the number of solutions while maintaining as much as possible the convergence and distribution of the \gls{pof}. 

We apply our proposed algorithm to the real-world Mission Planning Problem in which a team of \glspl{uav} must perform several tasks in a geodesic scenario in a specific time while being controlled by several \glspl{gcs}. In this context, there are several variables that influence the selection of the most appropriate plan, such as the makespan of the mission, the cost or the risk. Some experiments have been designed for evaluating the decrement of the number of solutions obtained, while maintaining the most significant ones.

This paper has been organized as follows. Section II provides some basics on \glspl{mop} and an introduction to the main concepts of Cone Domination. Section \ref{kpemo} presents the novel Knee-Point based Evolutionary Multi-Objective Optimization approach. In Section \ref{experimentalresults} this new algorithm is evaluated using a set of real Mission Planning Problems, and compared against our previous approach. Finally, Section V presents several conclusions that have been achieved from current work.

\section{Background}\label{background}
This section provides the background and related works concerning the cone-domination and metrics upon which is our proposed algorithm is based. 

\subsection{Multi-Objective Optimization}\label{mop}

In most \glspl{mop}, it is not possible to find one single optimal solution that could be selected as the best one; so in this kind of problems there is usually a set of solutions that represent several agreements between the given criteria. Any minimization \gls{mop} can be formally defined as:

\begin{align}
min \, \mathbf{f}(\mathbf{x})=(f_1(\mathbf{x}), f_2(\mathbf{x}), ..., f_m(\mathbf{x}))^T \nonumber \\
\text{subject to } \mathbf{x} \in \Omega \subseteq \mathbb{R}^n
\end{align}

\noindent 
where $\mathbf{x}=(x_1,x_2,...,x_n)^T$ represents a vector of $n$ decision variables, which are taken from the decision space $\Omega$; $\mathbf{f}:\Omega \rightarrow \Theta \subseteq \mathbb{R}^m$, where $f$ represents a set of $m$ objective functions, and it is possible to define a mapping from $n$-dimensional decision space $\Omega$ to $m$-dimensional objective space $\Theta$.

\begin{definition}{}
Given two decision vectors $\mathbf{x},\mathbf{y}\in \Omega$, $\mathbf{x}$ is said to \textbf{Pareto dominate} $\mathbf{y}$, denoted by $\mathbf{x} \prec \mathbf{y}$, iff:
\begin{align}
\forall i \in \{1,2,...,m\} \quad f_i(\mathbf{x}) \leq f_i(\mathbf{y}) \nonumber \\
\exists j \in \{1,2,...,m\} \quad f_j(\mathbf{x}) < f_j(\mathbf{y})
\end{align}
\end{definition}

\begin{definition}{}
A decision vector $\mathbf{x}^*\in \Omega$, is \textbf{Pareto optimal} if $\nexists \mathbf{x} \in \Omega$, $\mathbf{x} \prec \mathbf{x}^*$.
\end{definition}

\begin{definition}{}
The \textbf{Pareto set} $PS$, is defined as:
\begin{equation}
PS=\{\mathbf{x} \in \Omega|\mathbf{x} \text{ is Pareto optimal}\}
\end{equation}
\end{definition}

\begin{definition}{}
The \textbf{Pareto front} $PF$, is defined as:
\begin{equation}
PF=\{\mathbf{f}(\mathbf{x}) \in \mathbb{R}^m|\mathbf{x} \in PS\}
\end{equation}
\end{definition}

The goal of \gls{moea} is to find the non-dominated objective vectors which are as close as possible to the $PF$ (convergence) and evenly spread along the $PF$ (diversity). \gls{nsga2} has been one of the most popular algorithms over the last decade in this field \cite{DebEtAl2002}. This algorithm generates a non-dominated ranking in order to look for the convergence, and crowding  distance to assure the diversity of the solutions. Other popular algorithms are SPEA2 \cite{Zitzler2002}, MOEA/D \cite{Zhang2007} and NSGA-III \cite{Jain2013}. These last two algorithms have become very popular in the last years for their good performance on \glspl{maop}.

\subsection{Knee Points and Cone Domination}\label{kneepoints}
\begin{figure}[h]
    \centering
    \includegraphics[width=0.4\textwidth]{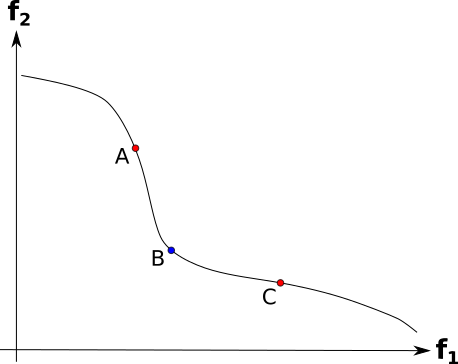}
    \caption{A synthetic simple example of a Pareto Front in a bi-objective minimization problem, where B is a knee point, while A and C, that have worse trade off than A, are not.}
    \label{fig:knee_points}
\end{figure}

In the last decade, several \glspl{moea} have been proposed to search for non-dominated solutions located close to a given reference point that incorporates the preference of the \gls{dmaker} \cite{PurshouseEtAl2014}. However, some a priori knowledge is require to set the reference point, which many times the \gls{dmaker} does not have. With the aim of obtaining significant solutions when no a priori knowledge is provided, the concept of finding "knee points" \cite{DebEtAl2003} can be used. In this way, we reduce the size of the non-dominated set and provide the \gls{dmaker} a small set of so-called knee point solutions. When distinct knee points are present in the Pareto front, most \glspl{dmaker} would prefer the solutions in these points, because if a near solution to a knee point (trying to improve slightly one objective) is selected, it will generate a large worsening at least in one of the other objectives. An example showing the difference between a knee point (blue) and points that are not knee points (red) is shown in Figure \ref{fig:knee_points}. 

The concept of using knee points has been studied before. Branke et al. \cite{Branke2004} presented a modification of \gls{nsga2} where the crowding distance criterion is computed using angle-based and utility-based measures for focusing on knee points. In Sch\"utze et al. \cite{Schutze2008} two different update methods are presented, these methods are based on maximal convex bulges that allow to focus the algorithm search on the knee points. Bechikh et al. \cite{Bechikh2011} extended the reference point \gls{nsga2} so the normal boundary intersection method is used to emphasize knee-like points. Zhang et al. \cite{Zhang2015} designed KnEA, an elitist Pareto-based algorithm that uses Knee point neighbouring as a secondary selection criteria in addition to dominance relationship.

In this paper, an angle-based measure is used to guide and focus the searching process on the knee points. Therefore, the concept of domination criterion has been changed to \textit{cone-domination}. The cone-domination has been defined as a weighted function that manages the set of objectives~\cite{DebEtAl2003}. This concept can be formally described as:

\begin{equation}\label{eq:cone-domination}
\Omega_i(\mathbf{f}(\mathbf{x}))=f_i(\mathbf{x}) + \sum_{j = 1,j \neq i}^{m} a_{ij}f_j(\mathbf{x}), \quad i=1,2,...,m
\end{equation}

\noindent 
where $a_{ij}$ is the amount of gain in the $j$-th objective function for a loss of one unit in the $i$-th objective function. The matrix $\mathbf{a}$, composed of these $a_{ij}$ values, and with 1 values in its diagonal elements, has to be provided in order to apply the above equations.

\begin{definition}{}
A solution $\mathbf{x}$ is said to \textbf{cone-dominate} a solution $\mathbf{y}$, denoted by $\mathbf{x} \prec^c \mathbf{y}$, if:
\begin{align}
\forall i \in \{1,2,...,m\} \quad \Omega_i(\mathbf{f}(\mathbf{x})) \leq \Omega_i(\mathbf{f}(\mathbf{y})) \nonumber \\
\exists j \in \{1,2,...,m\} \quad \Omega_j(\mathbf{f}(\mathbf{x})) < \Omega_j(\mathbf{f}(\mathbf{y}))
\end{align}
\end{definition}

In a bi-objective problem, the two ($m=2$) related objective weighted functions, can be defined as follows:

\begin{align}
\Omega_1(f_1,f_2)=f_1 + a_{12}f_2 \\
\Omega_2(f_1,f_2)=a_{21}f_1 + f_2
\end{align}

Previous equations can also be formalized as: 

\begin{equation}
\mathbf{\Omega} =
\begin{bmatrix}
	1 & a_{12}\\[0.3em]
    a_{21} & 1
\end{bmatrix}
\mathbf{f}, \quad \text{or, } \mathbf{\Omega} =\mathbf{a}\mathbf{f}
\end{equation}

\begin{figure}
    \centering
    \begin{subfigure}[b]{0.4\textwidth}
        \includegraphics[width=\textwidth]{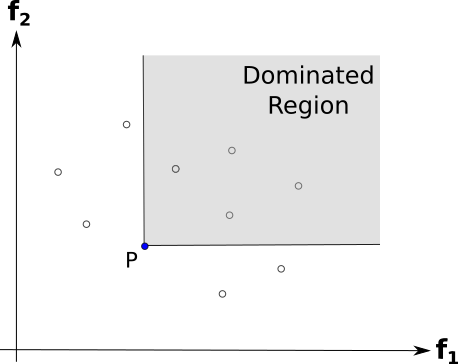}
        \caption{Pareto Dominated Region.}
        \label{fig:pareto_region}
    \end{subfigure}
    \qquad \qquad 
    \begin{subfigure}[b]{0.4\textwidth}
        \includegraphics[width=\textwidth]{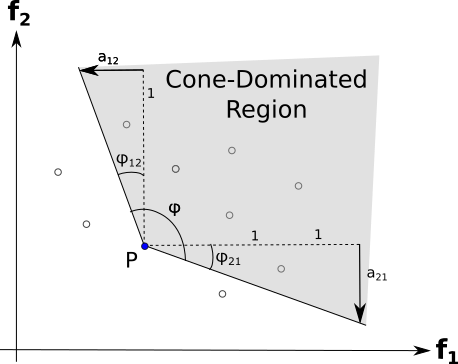}
        \caption{Cone Dominated Region.}
        \label{fig:cone_region}
    \end{subfigure}
    \caption{Regions dominated by a solution P (2.a) using the original definition of domination in a 2-objectives problem, and cone-dominated (2.b) by the same solution when the concept of cone domination is used.}\label{fig:domination}
\end{figure}

In Figure \ref{fig:cone_region} it is shown the contour lines for our previous two linear functions when these pass through a solution $P$ in objective space. The set of solutions inside those contour lines (the "cone-dominated region") will be dominated by $P$ according to the previous definition of domination. It is specially interesting to remark that when the standard definition of domination is used (see Figure \ref{fig:pareto_region}), the region between the horizontal and vertical lines will be dominated by $P$. Therefore, from both figures can be concluded that using the cone-domination definition will obtain larger regions (as the angle is greater than $90^{\circ}$), so more solutions will be dominated by one solution $P$ than when traditional definition is used. Therefore, and using the concept of cone-domination, the whole Pareto optimal front (using the traditional definition of domination), may not be non-dominated according to this new definition.

Besides, in Figure \ref{fig:cone_region} it can be observed that the values $a_{12}$ and $a_{21}$ expand the $\varphi$ angle modifying the dominated region (whose value in the original definition of Pareto dominance is $90^{\circ}$). In this example, the vertical axis is rotated by the angle of $\varphi_{12}$, whereas the horizontal axis is rotated by $\varphi_{21}$. As it is shown in this figure, both angles are related to the $a_{12}$ and $a_{21}$ values, respectively, as follows:

\begin{align}
tan \, \varphi_{12} = a_{12} \\
tan \, \varphi_{21} = a_{21}
\end{align}

Using previous equations, the new angle for the new dominated region of point $A$ will be $\varphi = 90^{\circ} + \varphi_{12} + \varphi_{21}$. If the values of the objectives are normalized, to make the dominated region symmetric and thus equalize the turn of both horizontal and vertical axes (i.e. $\varphi_{12} = \varphi_{21}$), both variables of the matrix must also be equalized: $a_{12} = a_{21}=tan\frac{\varphi-90}{2}$. Then, the cone-domination considers angles, ($\varphi_{12}, \varphi_{21}$) $\in$ [$90^{\circ}$, $180^{\circ}$]
, where the 90 degrees case is the common Pareto dominance, and the 180 degrees is equivalent to a weighted sum multi-objective optimization where all weights are the same (i.e. a single-objective optimization using the sum of all objectives as fitness function). In this last case, all the solutions are inside the same line of cone-domination, and the matrix $\mathbf{a}$ considered is filled with 1 ($\forall i,j \quad a_{ij}=1$).


In other cases, the cone domination concept can be formally defined as:

\begin{equation}
\mathbf{\Omega} =
\begin{bmatrix}
	1 & tan\frac{\varphi-90}{2}\\[0.3em]
    tan\frac{\varphi-90}{2} & 1
\end{bmatrix}
\mathbf{f}
\end{equation}

With the aim of comparing the convergence and diversity of the knee points obtained, in contrast with the Pareto front from the original approach, a study of several values of the angle $\varphi \in (90,180)$ must be carried out.

Now, let extend this concept to higher dimensions. For three dimension ($m=3$), in this case, having $\mathbf{f}=(f_1,f_2,f_3)^T$, the cone domination function $\mathbf{\Omega} =\mathbf{a}\mathbf{f}$ is expressed as:

\begin{equation}
\mathbf{\Omega} =
\begin{bmatrix}
	1 & a_{12} & a_{13}\\[0.3em]
    a_{21} & 1 & a_{23}\\[0.3em]
    a_{31} & a_{32} & 1
\end{bmatrix}
\mathbf{f}
\end{equation}

\begin{figure*}[!t]
    \centering
    \begin{subfigure}[b]{0.4\textwidth}
        \includegraphics[width=\textwidth]{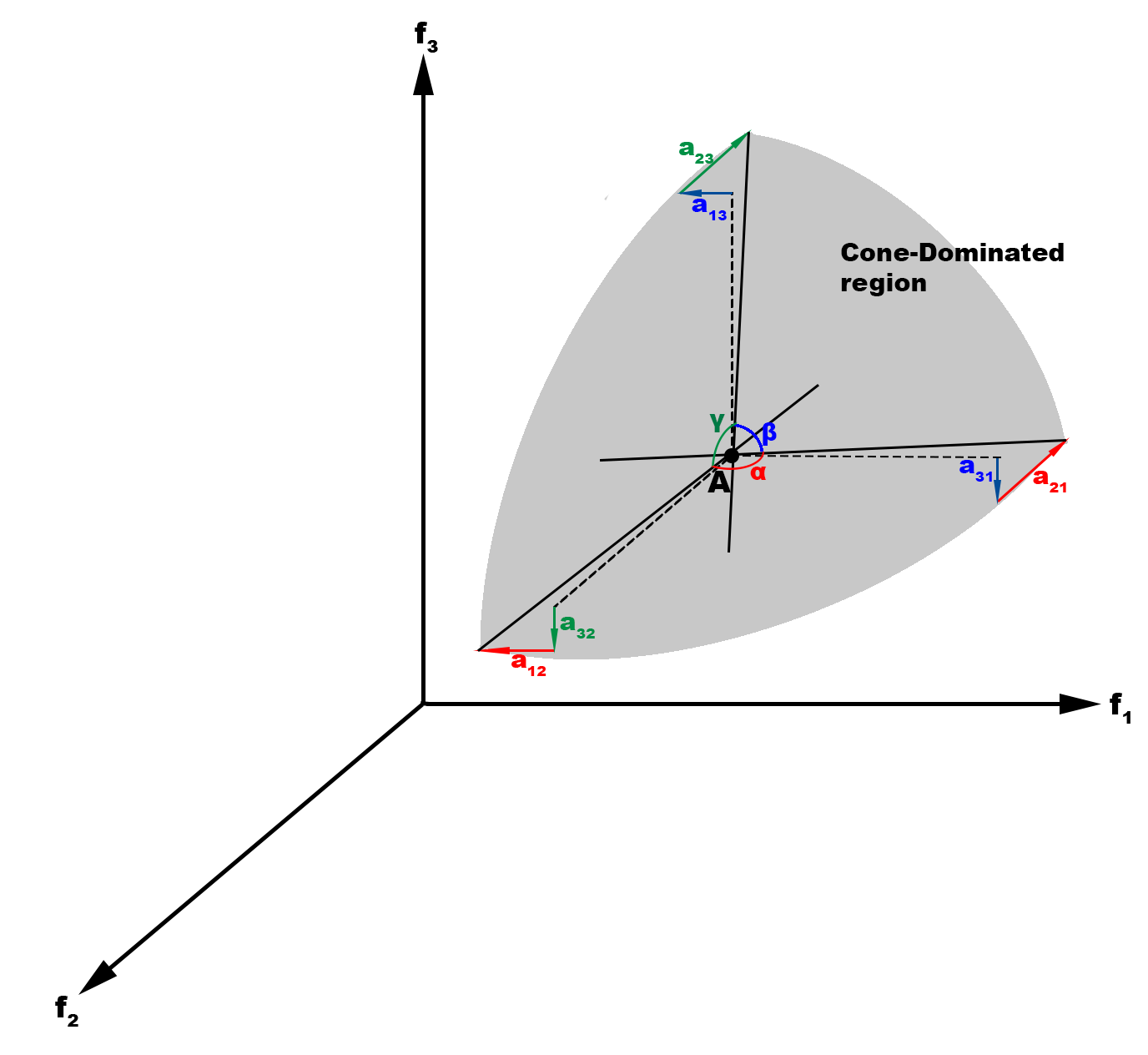}
        \caption{Cone Dominated Region.}
        \label{fig:cone_region3D}
    \end{subfigure}
    \qquad \qquad 
    \begin{subfigure}[b]{0.4\textwidth}
        \includegraphics[width=\textwidth]{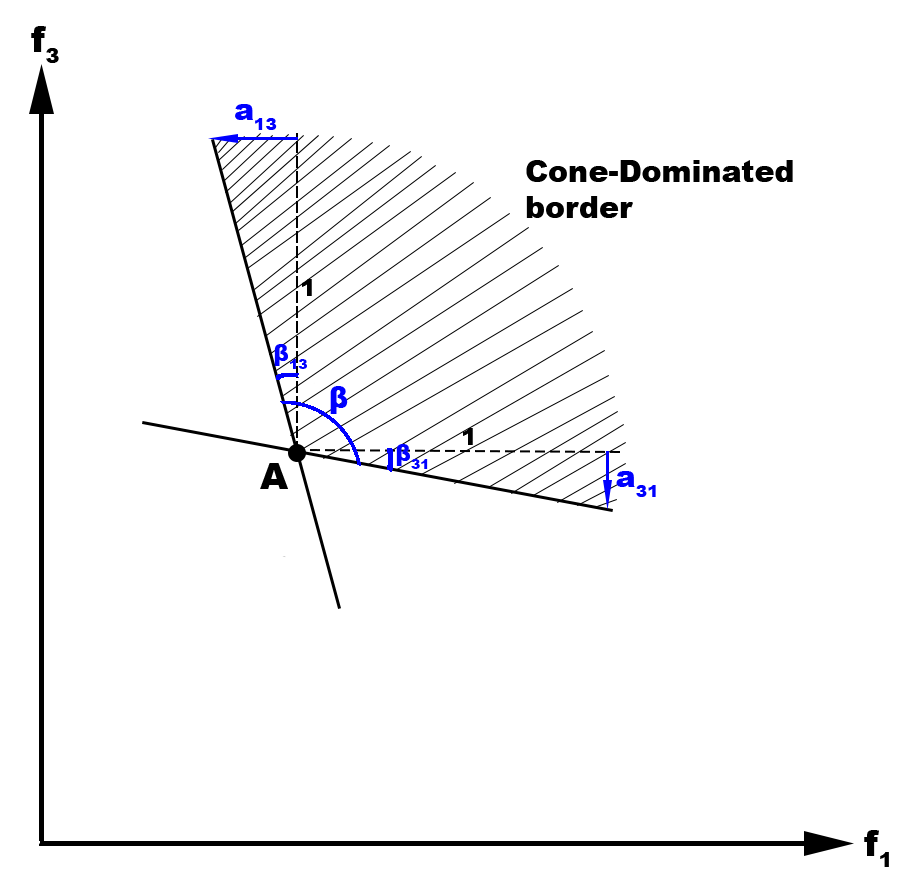}
        \caption{Projection in the f1-f2 plane.}
        \label{fig:cone_region3D_projXY}
    \end{subfigure}
    \qquad \qquad 
    \begin{subfigure}[b]{0.4\textwidth}
        \includegraphics[width=\textwidth]{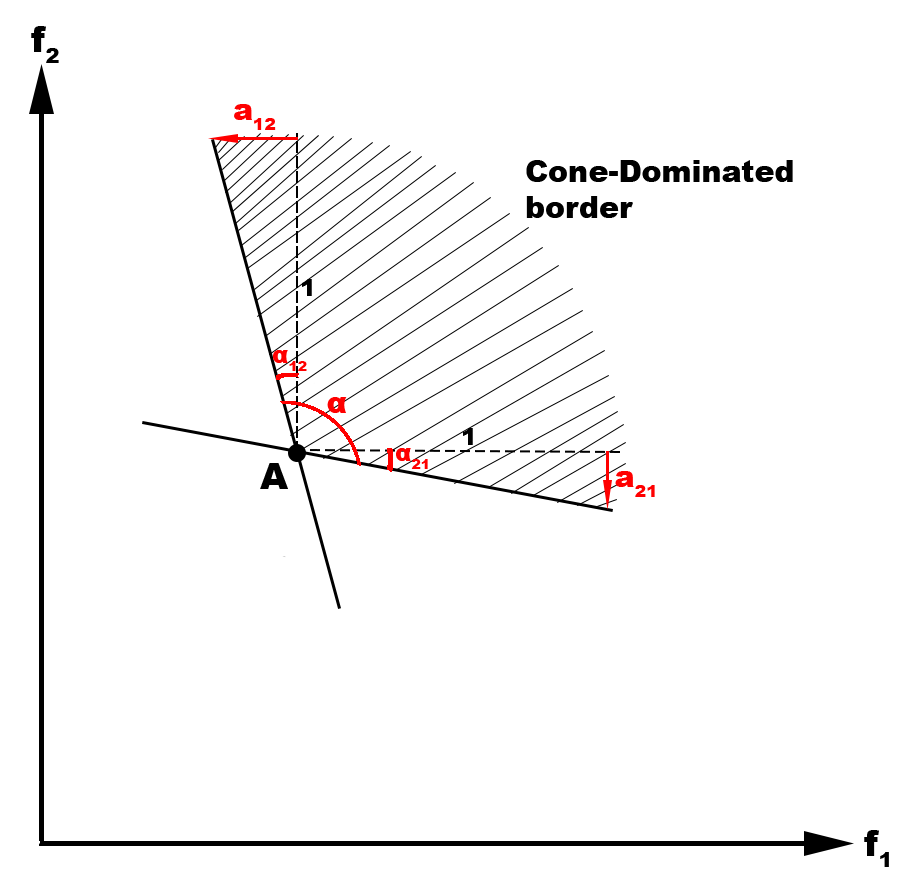}
        \caption{Projection in the f1-f3 plane.}
        \label{fig:cone_region3D_projXZ}
    \end{subfigure}
    \qquad \qquad 
    \begin{subfigure}[b]{0.4\textwidth}
        \includegraphics[width=\textwidth]{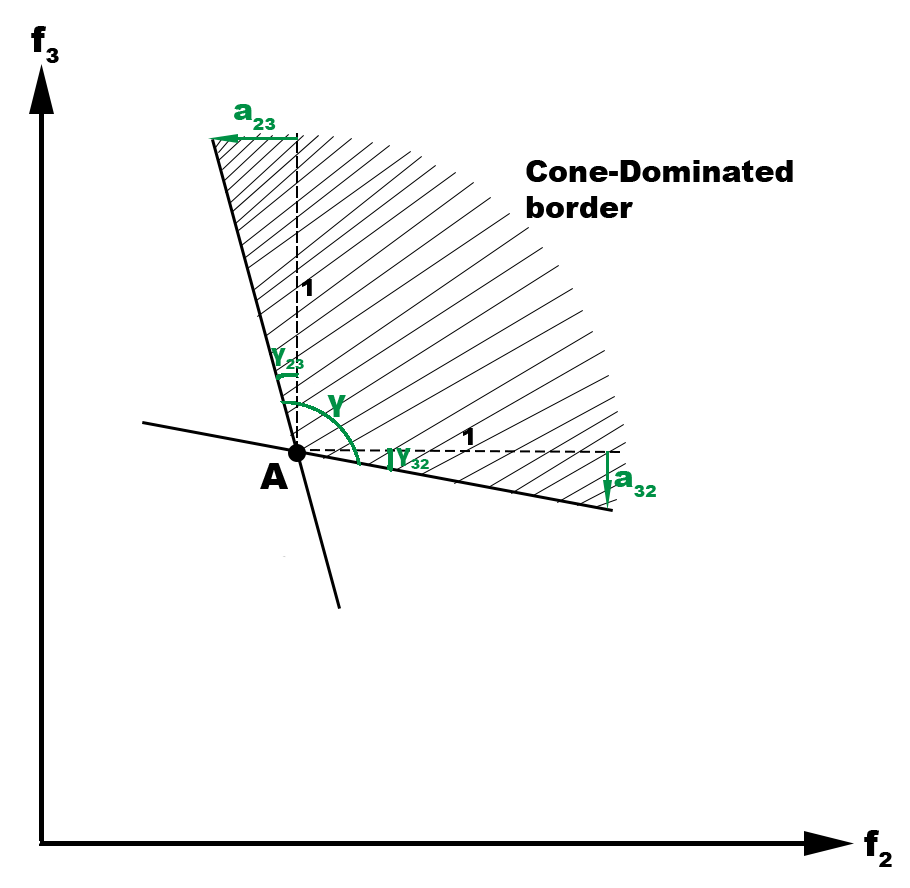}
        \caption{Projection in the f2-f3 plane.}
        \label{fig:cone_region3D_projYZ}
    \end{subfigure}
    \caption{Regions cone-dominated by solution A in a 3-objective problem}\label{fig:domination3D}
\end{figure*}

Figure \ref{fig:cone_region3D} shows the 3D contour corresponding to cone-dominated region for a solution $A$ in the objective space, where the bold lines converging in $A$ represent the edges of the cone region. In dashed lines, the edges used in the normal definition of Pareto domination for $A$ are also presented. As in the 2D case, the modified definition of domination allows a larger region to become dominated by any solution than the usual definition.

Besides, Figures  \ref{fig:cone_region3D_projXY}, \ref{fig:cone_region3D_projXZ} and \ref{fig:cone_region3D_projYZ} show the projections of this 3D region into the $f_1$-$f_2$ plane, $f_1$-$f_3$ plane and $f_2$-$f_3$ plane, respectively. These projections show a similarity with the cone-domination region for 2-objectives problems (see Figure \ref{fig:cone_region}). In Figure \ref{fig:cone_region3D_projXY} it is observable that the values $a_{12}$ and $a_{21}$ change the dominated region by expanding the $\alpha$ value, rotating the f1 axis an angle of $\alpha_{21}$ and the f2 axis an angle of $\alpha_{12}$. On the other hand, Figure \ref{fig:cone_region3D_projXZ} shows how values $a_{13}$ and $a_{31}$ expand the $\beta$ angle, rotating the f1 axis an angle of $\beta_{31}$ and the f3 axis an angle of $\beta_{13}$. Finally, Figure \ref{fig:cone_region3D_projYZ} shows how values $a_{23}$ and $a_{32}$ expand the $\gamma$ angle, rotating the f2 axis an angle of $\gamma_{32}$ and the f3 axis an angle of $\gamma_{23}$. As can be seen, similarly to the 2D cone-domination, the angles described and values of matrix $\mathbf{a}$ can be formulated as follows:

\begin{align}
tan \, \alpha_{12} = a_{12}, \quad  tan \, \alpha_{21} = a_{21}, \nonumber \\
\alpha = 90 + \alpha_{12} + \alpha_{21}
\end{align}

\begin{align}
tan \, \beta_{13} = a_{13}, \quad  tan \, \beta_{31} = a_{31}, \nonumber  \\
\beta = 90 + \beta_{13} + \beta_{31}
\end{align}

\begin{align}
tan \, \gamma_{23} = a_{23}, \quad  tan \, \gamma_{32} = a_{32}, \nonumber  \\
\gamma = 90 + \gamma_{23} + \gamma_{32}
\end{align}

Following this reasoning, the cone-dominated region is defined by extending the angle of the region in 2-objective problems, while in 3-objective problems it is defined by extending the three angles of the faces defining the 3D cone. When considering a higher dimension $N$, a hypercone region with $\begin{pmatrix}
	N \\
    2
\end{pmatrix}$ faces (and therefore, angles) is generated. This number is the result of the pair-combinations of the N objective functions. In this case, the cone domination is expressed as:

\begin{equation}
\mathbf{\Omega} =
\begin{bmatrix}
	1 & a_{12} & \dots & a_{1n}\\[0.3em]
    a_{21} & 1 & \dots & a_{2n}\\[0.3em]
    \vdots & \vdots & \ddots  & \vdots \\[0.3em]
    a_{n1} & a_{n2} & \dots &  1
\end{bmatrix}
\mathbf{f}
\end{equation}

\noindent
where each value $a_{ij}$ with $i \neq j, \; i,j \in \{1,2,...,N\}$ is related to the angle $\theta_{ij}$ so $tan \, \theta_{ij} = a_{ij}$. Besides, the angle $\theta^{ij}$ of the face $f_i$-$f_j$ of the hypercone region, is related to this angle (hence, $\theta^{ij}$ can be calculated as $\theta^{ij} = 90 + \theta_{ij} + \theta_{ji}$).

Again, and in order to make the cone dominated region symmetric for every objective, and supposing that the values of the objectives are normalized, it is necessary to equalize the angles of every face of the hypercone region. The definition of this angle $\theta$ leads to the setting of the values $a_{ij}$ of matrix $\mathbf{a}$:

\begin{equation}\label{eq:simetric-cones}
a_{ij} = tan\frac{\theta-90}{2}, \; \forall i,j \in \{1,2,...,N\},\; i \neq j
\end{equation}

\subsection{Hypervolume and distribution of solutions}\label{hypervolumedistribution}

In order to evaluate the convergence and distribution of the non-dominated solutions obtained with \glspl{moea}, some metrics have been proposed over the last decades \cite{Jiang2014}. One of the most popular is the \gls{hv} \cite{Fonseca2006}, which gives the volume (in the objective space) that is dominated by some reference point. Other frequently used metric is the \gls{igd} \cite{Ishibuchi2015}, where a set of reference points is provided as an approximation of the Pareto front, and the \gls{igd} is computed as the distance from each reference point to the nearest solution in the solution set.

When the number of non-dominated solutions is large, the decision making process afterwards becomes really complex. In order to avoid this, the best outcome of the algorithm should be a small set of solutions maintaining as large as possible value for the \gls{hv} or \gls{igd}. As the \gls{igd} is pretty complex to compute due to the need of a reference set, which in real problem is sometimes difficult to provide, the \gls{hv} metric has been used to design a new metric that takes into account the number of solutions. We propose the hypervolume-distribution (HDist) metric, which is designed in order to evaluate this trade-off between \gls{hv} and the number of solutions. This metric is defined as follows:

\begin{align}
HDist(S,P) = \frac{\sharp(P)-\sharp(S)}{\sharp(P)} \times \frac{HV(S)}{HV(P)}
\end{align}

where $P$ is the \gls{pof} of the problem, $S \subset P$ is the set of non-dominated solutions to be evaluated, and $HV : \mathbb{R}^n \rightarrow \mathbb{R}$ represents the hypervolume of the set. In this metric, the Pareto set is needed in order to normalize the number of solutions and the hypervolume, which are then combined. The higher the value of this metric, the better distributed solutions maintaining good hypervolume.

Using this metric, our main goal is to self-adapt the cone-domination angle in the evolving phase of the \gls{moea}, so that the best set of non-dominated solutions according to the HDist metric is obtained.

\section{sKPNSGA-II: a self-adaptive Knee-point based extension of NSGA-II}
\label{kpemo}

In order to reduce the number of obtained solutions in a \gls{mop}, we propose an extension of \gls{nsga2}, which is designed to search Knee-points instead of non-dominated solutions. To reach this goal, the cone-domination concept described in section \ref{kneepoints} will be used instead of the standard Pareto-domination. The non-dominated ranking used by \gls{nsga2} is changed to a non-cone-domination ranking with an specific angle $\theta$. In a previous work \cite{Ramirez-Atencia2017}, a first approach called \gls{kpnsga2}, using cone-domination with a fixed cone angle, was proposed. In this work, we propose \gls{skpnsga2}, which self-adapts the cone angle according to the $HDist$ metric proposed in previous section. In order to perform this self-adaptation, the golden section search \cite{Press2007} has been used. This technique is used to find the maximum value for the $HDist$ metric through a successive narrowing of the the range of values in which the maximum point is located.

The \gls{skpnsga2} is presented in Algorithm \ref{alg:kpnsga2}. This novel approach, after randomly generating the initial population (Line 1), initializes the convergence factors (Lines 2-4). Following this step, the maximum and minimum values for each objective is initialized to a zero-vector $q$ (line 5), and to the vector $\overline{M}$ of maximum objective values (Line 6), respectively. Every time the solutions are evaluated, these values are updated. Their aim is to be used in the normalization of the objective values.

\begin{algorithm}[!h]
 \caption{Self-adaptive Knee-Point based NSGA-II.}
 \label{alg:kpnsga2}
\DontPrintSemicolon
\KwIn{ A problem $P$. The set of $m$-objectives $O$ and their upper bounds $\overline{M}=\{M_i >> avg(o_i)\}$. And a set of positive parameters: $\mu$ (elitism),  $\lambda$ (population size), $mutprobability$, $stopGen$ (stopping criteria limit), and $maxGen$ (maximum number of generations). $\phi = \frac{\sqrt{5}+1}{2}$ is the golden ratio.}
\KwOut{The Knee-Point Frontier generated}
$S \gets$ set of $\lambda$ individuals randomly generated \;
 $i \gets 1$\;
 $convergence \gets 0$ \;
 $kpof \gets \emptyset$ \;
 $maxP \gets [0,...,0]^q$\;
 $minP \gets \overline{M}$ \;
 $\theta,\theta_A,\theta_C \gets 90$\;
 $\theta_B,\theta_D \gets 180$\;
 \While{$i \leq maxGen \land convergence < stopGen$}{
  	\For{$j \gets 1$ \textbf{to} $|S|$}{
    	$f.objectives \gets MultiObjectiveFitness(S_j, O)$\;
        $maxP \gets maxPerElem(maxP,f.objectives)$\;
        $minP \gets minPerElem(minP,f.objectives)$\;
        $S_j.fit \gets f$ \;
    }
    $S \gets buildArchive(S, \lambda, \theta, maxP, minP)$ \;
    $newkpof \gets kneeFront(S,\theta, maxP, minP)$ \;
    \If {$newkpof = pof$}{  
    	$convergence \gets convergence + 1$ \;
    }
    $goldenSection(newkpof,\theta,\theta_A,\theta_B,\theta_C,\theta_D)$\;
    $kpof \gets newkpof$\;
    $newS \gets SelectElites(S, \mu)$ \;
    \For{$j \gets \mu$ \textbf{to} $\lambda$}{
      $p1,p2 \gets TournamentSelection(S$) \;
      $i1,i2 \gets Crossover(p1, p2)$ \;
      $i1 \gets Mutation(i1,mutprobability)$ \;
      $i2 \gets Mutation(i2,mutprobability)$ \;
      $newS \gets newS \cup \{i1, i2\}$ \;
    }
    $S \gets S \cup newS$ \;
 }
 \Return{kpof}\;
\end{algorithm}

The fitness function (Lines 10-14) used in the evaluation of the individuals computes the multi-objective values of the solutions, which are stored inside the fitness. Moreover, as previously mentioned, the maximum and minimum objective values are updated with the new evaluated solutions (Lines 12-13).

Based on the \gls{nsga2} algorithm, the new offspring is updated with the \emph{buildArchive} function (Algorithm \ref{alg:kpnsga2}, line 15), which is presented in Algorithm \ref{alg:build-archive}. This function creates an array of vectors, or fronts, storing the solutions grouped by their level of non-cone-dominance. This is done using the \emph{assignFrontRanks} function (see Algorithm \ref{alg:front-ranks}). In this levelled array, the first front is composed of the non-cone-dominated solutions of the population; the second front contains the non-cone-dominated solutions among the rest of the population without considering the solutions of the first front; the third front is then composed of the non-cone-dominated solutions of the population without considering the solutions of the first and second fronts, and so on.

\begin{algorithm}[!t]
 \caption{BuildArchive($S, \lambda, \theta, maxP, minP$)}
 \label{alg:build-archive}
\DontPrintSemicolon
\KwIn{ Vector $S$ containing the actual population. Population size $\lambda$. Angle $\theta$ for every face of the cone. Vector $maxP$ stores the maximum values found for the $q$ objectives. Vector $minP$ stores the minimum values found for all $q$ objectives.}
\KwOut{Updated vector $S$}
$newS \gets \emptyset$ \;
$ranks \gets assignFrontRanks(S, \theta, maxP, minP)$ \;
 \For{$i \gets 1$ \textbf{to} $|ranks|$}{
  	$rank \gets ranks[i]$ \;
    $assignSparsity(rank, maxP, minP)$ \;
  	\If {$|rank|+|newS| \geq \lambda$}{
    	$rank \gets sort(rank)$ \;
        $newS \gets newS \cup subVector(rank, 0, \lambda - |newS|)$ \;
    }
    \Else {
      $newS \gets newS \cup rank$ \;
    }
 }
 \Return{newS}\;
\end{algorithm}

\begin{algorithm}[!t]
 \caption{assignFrontRanks($S, \theta, maxP, minP$)}
 \label{alg:front-ranks}
\DontPrintSemicolon
\KwIn{ Vector $S$ containing the actual population. Angle $\theta$ for every face of the cone. Vector $maxP$ stores the maximum values found for the $q$ objectives. Vector $minP$ stores the minimum values found for all $q$ objectives.}
\KwOut{List of vectors containing the solutions with same rank values.}
$rankedFronts \gets \emptyset$ \;
$inds \gets S$ \;
$rank \gets 1$ \;
\While{$|inds| > 0$}{
  $front \gets kneeFront(inds,\theta, maxP, minP)$ \;
  $inds \gets inds - front$ \;
  \For{$j \gets 1$ \textbf{to} $|front|$}{
  	$front[j].fitness.rank \gets rank$ \;
  }
  $rankedFronts \gets rankedFronts \cup \{front\}$ \;
  $rank \gets rank + 1 $\;
}
 \Return{$rankedFronts$}\;
\end{algorithm}

In order to create the array of ranked fronts, the \emph{kneeFront} function is used (see Algorithm \ref{alg:knee-front}). This function is similar to the classical one used in \gls{nsga2} to generate the Pareto front from a population. Nevertheless, it has been changed, so instead of the non-dominated solutions, the function will consider the non-cone-dominated solutions, as was described in Section \ref{kneepoints}. The new approach of cone domination is described in detail in Algorithm \ref{alg:cone-dominate}. Then, \emph{kneeFront} function requires a $\theta$ value indicating the angle of the cone-domination. First, the objective vectors are normalized with the maximum and minimum values. Then, the cone-domination function is computed using the Equations \ref{eq:cone-domination} and \ref{eq:simetric-cones} for each objective; and the function examines if the second solution is cone-dominated by the first.

\begin{algorithm}[!t]
 \caption{kneeFront($S, \theta, maxP, minP$)}
 \label{alg:knee-front}
\DontPrintSemicolon
\KwIn{ Vector $S$ containing the actual population. Angle $\theta$ for every face of the cone. Vector $maxP$ stores the maximum values found for the $q$ objectives. Vector $minP$ stores the minimum values found for all $q$ objectives.}
\KwOut{Knee-Point Frontier based on the Cone-Domination with angle $\theta$.}
$front \gets \{S[1]\}$ \;
\For{$i \gets 2$ \textbf{to} $|S|$}{
  $noOneWasBetter \gets TRUE$ \;
  \For{$j \gets 1$ \textbf{to} $|front|$}{  \If{$ConeDom(S[j], S[i], \theta, maxP, minP)$}{
    	$noOneWasBetter \gets FALSE$ \;
    }
    \ElseIf{$ConeDom(S[i], S[j], \theta, maxP, minP)$} {
      $front \gets front - S[j]$ \;
      $j \gets j - 1$ \;
    }
  }
  \If{$noOneWasBetter$}{
  	$front \gets front \cup S[i] $ \;
  }
}
 \Return{$front$}\;
\end{algorithm}

\begin{algorithm}[!t]
 \caption{ConeDom($A, B,\theta, maxP, minP$)}
 \label{alg:cone-dominate}
 \DontPrintSemicolon
 \KwIn{Solutions A and B, used to check for cone-domination. The angle $\theta$ (in degrees) for every face of the cone. Vectors $maxP$ and $minP$ store the maximum and minimum values found for all $m$ objectives.}
 \KwOut{TRUE if A dominates B, FALSE otherwise.}
 $\mathbf{x} \gets \frac{A.fit.objectives - minP}{maxP - minP}$ \;
 $\mathbf{y} \gets \frac{B.fit.objectives - minP}{maxP - minP}$ \;
 $dominates \gets$ FALSE \;
 \For{$i \gets 1$ \textbf{to} $m$}{
   $cone1 \gets \mathbf{x}[i]$\;
   $cone2 \gets \mathbf{y}[i]$\;
   \For{$j \gets 1$ \textbf{to} $m$}{
     $cone1 \gets cone1 + tan(\frac{\theta-90}{2}) \cdot \mathbf{x}[j]$ \;
     $cone2 \gets cone2 + tan(\frac{\theta-90}{2}) \cdot \mathbf{y}[j]$ \;
   }
   \If {$cone1 < cone2$}{
   	 $dominates \gets$ TRUE \;
   }
   \If {$cone1 > cone2$}{
  	 \Return{FALSE}\;
   }
 }
 \Return{$dominates$}\;
\end{algorithm}

Once the array of vectors containing the ranked solutions is created, in a similar way to \gls{nsga2} algorithm, a sparsity value (that is based on the crowding distance) is given to each solution at every vector, through the \emph{assignSparsity} function in Algorithm \ref{alg:build-archive}.

\begin{figure}[!h]
		\includegraphics[width=0.45\textwidth]{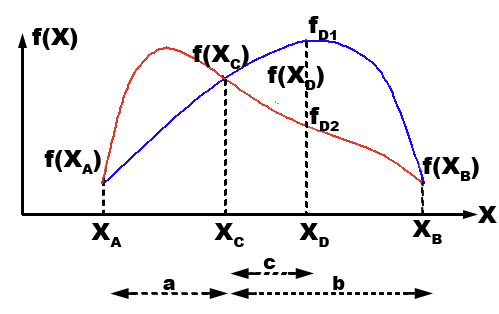}
		\centering
		\caption{Diagram of the Golden Section Search.}
		\label{fig:golden-section}
	\end{figure}

In order to self-adapt the cone angle according to the $HDist$ metric, the Golden Section Search has been used (Line 19). This technique is used to find the maximum of the $HDist$ metric iteratively as the main algorithm evolves. It is described in Algorithm \ref{alg:golden-section}. This technique is similar to the bisection search for the root of an equation. Specifically, if in the neighbourhood of the maximum we can find three points $x_A <x_C <x_B$ corresponding to $f(x_A) > f(x_C)< f(x_B)$, then there exists a maximum between the points $x_A$ and $x_B$. To search for this maximum, we can choose another point $x_D$ between $x_C$ and $x_B$ as shown in the figure \ref{fig:golden-section}. Then, depending on the value of $f(X_D)$, the new triplet may become $x_A <x_C <x_D$ if $f(X_D) = f_{D2}  < f(X_C)$, or $x_C <x_D <x_B$ if $f(X_D) = f_{D1} > f(X_C)$. And so, the process is repeated iteratively until an error tolerance is reached. In order to compute these points $x_C$ and $x_D$, the golden ratio $\phi$ is used, where each point is separated from the corner points $x_A$ and $x_B$ the distance between these corner points divided by $\phi$.

In \gls{skpnsga2}, the golden section search starts working once the front has a large number of solutions or the hypervolume does not show a considerable increase with respect to previous generations. Then, the $\theta_C$ cone angle value is tested in the following generation, and then the $\theta_D$ in the next one. After testing both, they are compared as previously described, the triplet is updated and the process continues until the stopping criteria is met.

\begin{algorithm}[!t]
 \caption{goldenSection($S, \theta, \theta_A, \theta_B, \theta_C, \theta_D$)}
 \label{alg:golden-section}
\DontPrintSemicolon
\KwIn{ Vector $S$ containing the actual knee front. Angle $\theta$ being used. Angles used in the golden section search $\theta_A$, $\theta_B$, $\theta_C$ and $\theta_D$.}
$hyp \gets HV(newkpof)$\;
$minHyp \gets min(hyp,minHyp)$\;
$maxHyp \gets max(hyp,maxHyp)$\;
$minPOF \gets min(|newkpof|,minPOF)$\;
$maxPOF \gets max(|newkpof|,maxPOF)$\;
\If{$\theta = 90$}{
	\If{$|newkpof| > \mu \quad \lor$ $ HV(newkpof)-HV(kpof) < 10^{-5}$}{
		$\theta_C \gets \theta_B - \frac{\theta_B-\theta_A}{\phi}$\;
		$\theta_D \gets \theta_A + \frac{\theta_B-\theta_A}{\phi}$\;
		$\theta \gets \theta_C$\;
		$test_C \gets true$\;
	}
}
\Else {
	\If{$test_C$}{
		$HDist_C \gets \frac{hyp-minHyp}{maxHyp-minHyp} \times \frac{maxPOF-|newkpof|}{maxPOF-minPOF}$\;
		$\theta \gets \theta_D$\;
		$test_C \gets false$\;
	}
	\Else{
		$HDist_D \gets \frac{hyp-minHyp}{maxHyp-minHyp} \times \frac{maxPOF-|newkpof|}{maxPOF-minPOF}$\;
		\If{$HDist_C > HDist_D$}{
			$\theta_B \gets \theta_D$\;
		}
		\Else{
			$\theta_A \gets \theta_C$\;
		}
		$\theta_C \gets \theta_B - \frac{\theta_B-\theta_A}{\phi}$\;
		$\theta_D \gets \theta_A + \frac{\theta_B-\theta_A}{\phi}$\;
		$\theta \gets \theta_C$\;
		$test_C \gets true$\;
	}
}
\end{algorithm}

Following Algorithm \ref{alg:kpnsga2}, a tournament selection (Line 23) is used to provide the individuals that will be chosen to apply the genetic operators. The crossover operator (line 24) and the mutation operator (lines 25-26) are then applied.

Finally, in this algorithm the stopping criteria considers the comparison of the non dominated solutions obtained so far at the end of each generation with the solutions from the previous generation (Lines 17-18). When the solutions obtained so far remain unchanged for a specific number of generations, the algorithm will stop and return the set of solutions found as the best approximation of the \gls{pof}.

\section{Experimental evaluation}\label{experiments}

\label{experimentalsetup}

In order to test the proposed algorithm, a real complex problem have to be considered where decision makers actually care about the number of solutions for the decision making process. In these experiments, several real Mission Planning Problems have been designed for this. Mission Planning\cite{Ramirez-Atencia2018Constrained} is a complex problem that involves the assignment of several tasks to the vehicles performing them, along with the assignments of vehicles to \glspl{gcs} controlling them. Some tasks are performed by just one vehicle, while others may be performed by several vehicles reducing the time needed for the task (e.g. taking a photo, monitoring a target\ldots). There exists several issues to take into account, such as the paths followed by the \glspl{uav} when there are \glspl{nfz} in the scenario, the sensors to be used by the vehicles for each task, the flight time or the fuel consumption, among others. In a previous work\cite{Ramirez-Atencia2018Weighted}, this problem was modelled as a \gls{csp}, considering the different constraints of the problem (sensors, path, time, fuel\ldots), and solved using a standard \gls{nsga2} algorithm.


This problem is also a Multi-Objective Optimization problem, as there exist several objectives that influence the selection of the most appropriate plan. \textbf{7 objectives} have been considered, that include: the \textbf{total cost} of the vehicles for completing the mission; the \textbf{makespan} or end time when all vehicles have returned and the mission is ended; or the \textbf{risk} of the mission, which has been calculated as an average percentage that indicates how hazardous the mission is (e.g. \glspl{uav} that end up with low fuel, \glspl{uav} that fly near to the ground or \glspl{uav} that fly close between them); the \textbf{number of \glspl{uav}} used in the mission, the \textbf{total fuel consumption}, the \textbf{total flight time} and the \textbf{total distance traversed}.

The fitness function used for this problem checks that all of the constraints considered are fulfilled for a given solution. If not, it stores inside its fitness the number of constraints fulfilled by the solution. When all constraints are fulfilled, the fitness will work as a multi-objective function minimizing the problem objectives.

The encoding considered here takes into account the different variables of the \gls{csp} model, which includes: the assignments of \glspl{uav} to tasks, the order of the tasks, the assignments of \glspl{gcs} to \glspl{uav}, the flight profiles used in each path and return to the base, and the sensors used for each task performance. For this encoding, proper crossover and mutation operators have been designed, where a concrete operator is applied to each allele of the individuals. For more details about the encoding and the \gls{csp} model, may you consult previous works \cite{Ramirez-Atencia2018Constrained} \cite{Ramirez-Atencia2018Weighted}.

\begin{figure*}[!t]
    \centering
    \begin{subfigure}[b]{0.23\textwidth}
        \includegraphics[width=\textwidth]{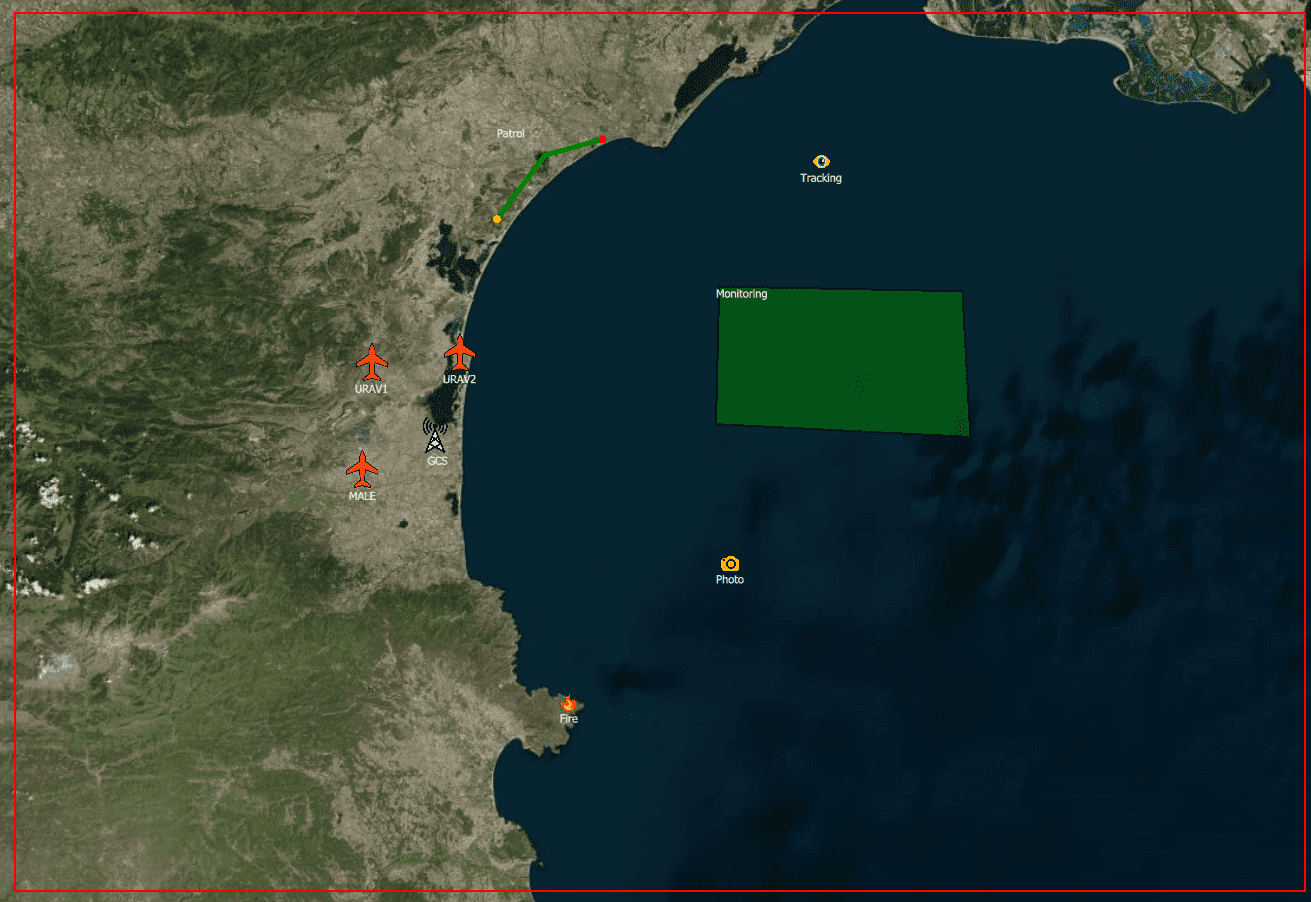}
        \caption{Mission 1.}
        \label{fig:d11}
    \end{subfigure}
    \quad
    \begin{subfigure}[b]{0.23\textwidth}
        \includegraphics[width=\textwidth]{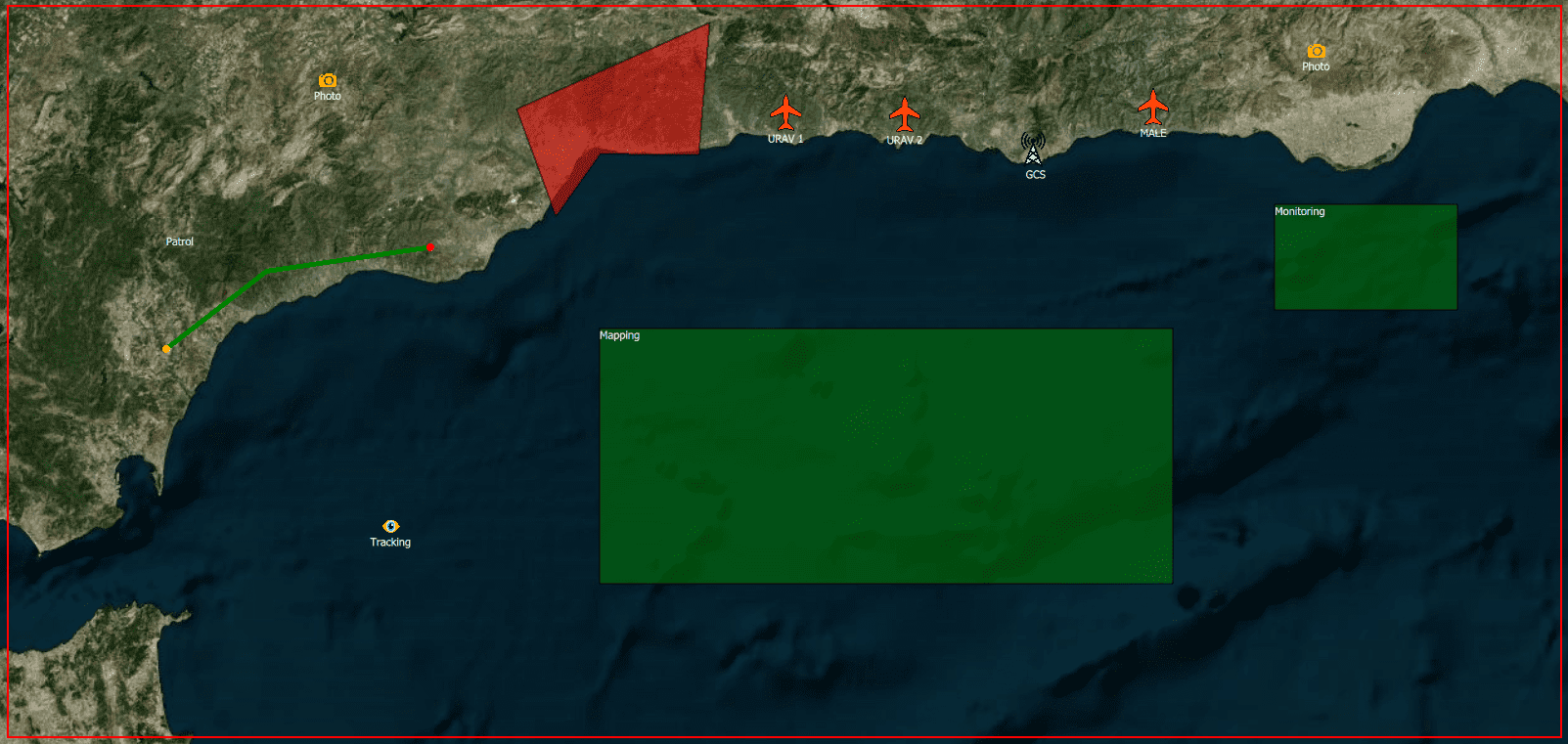}
        \caption{Mission 2.}
        \label{fig:d12}
    \end{subfigure}
    \quad
    \begin{subfigure}[b]{0.23\textwidth}
        \includegraphics[width=\textwidth]{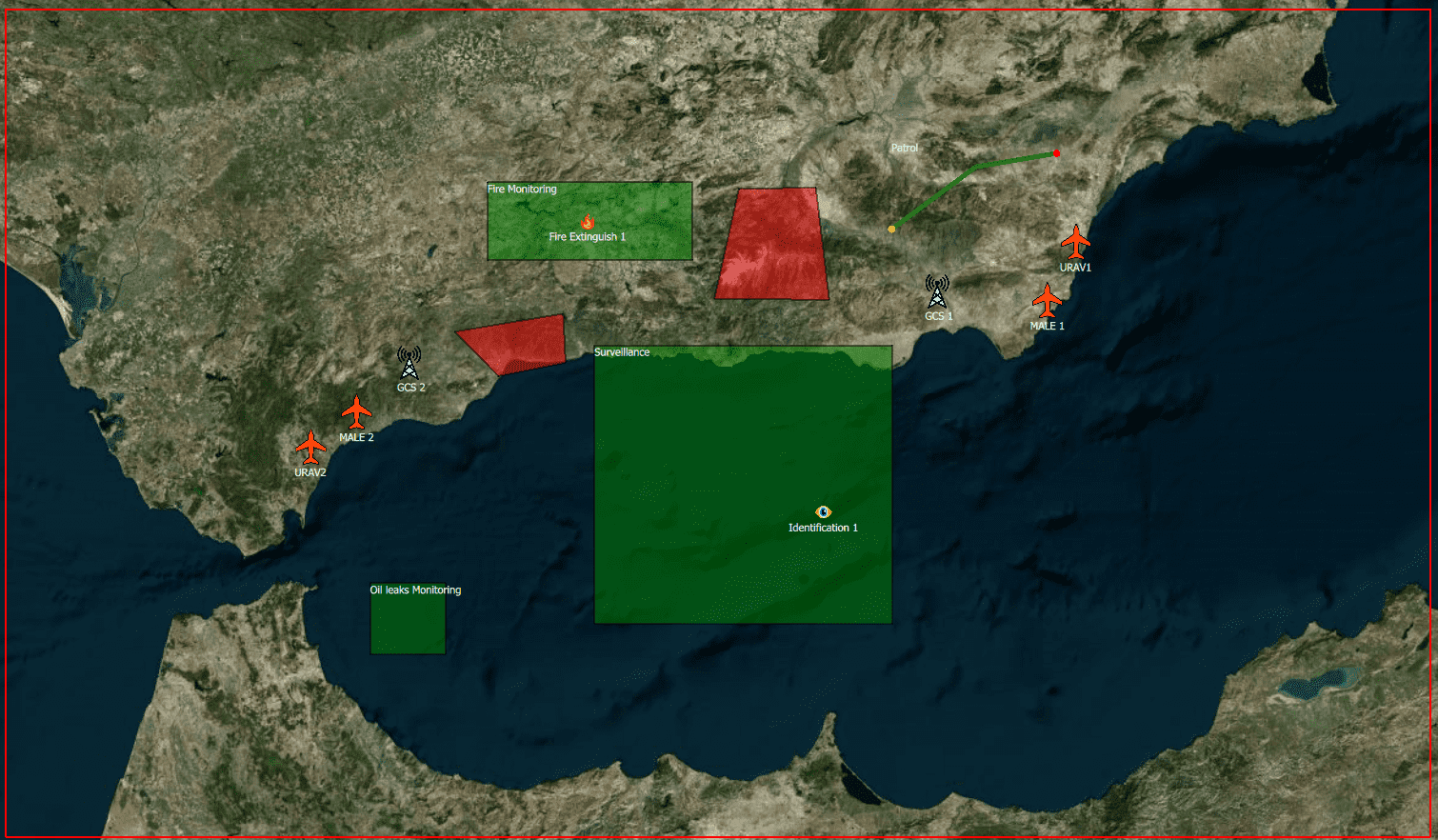}
        \caption{Mission 3.}
        \label{fig:d13}
    \end{subfigure}
    \quad
    \begin{subfigure}[b]{0.23\textwidth}
        \includegraphics[width=\textwidth]{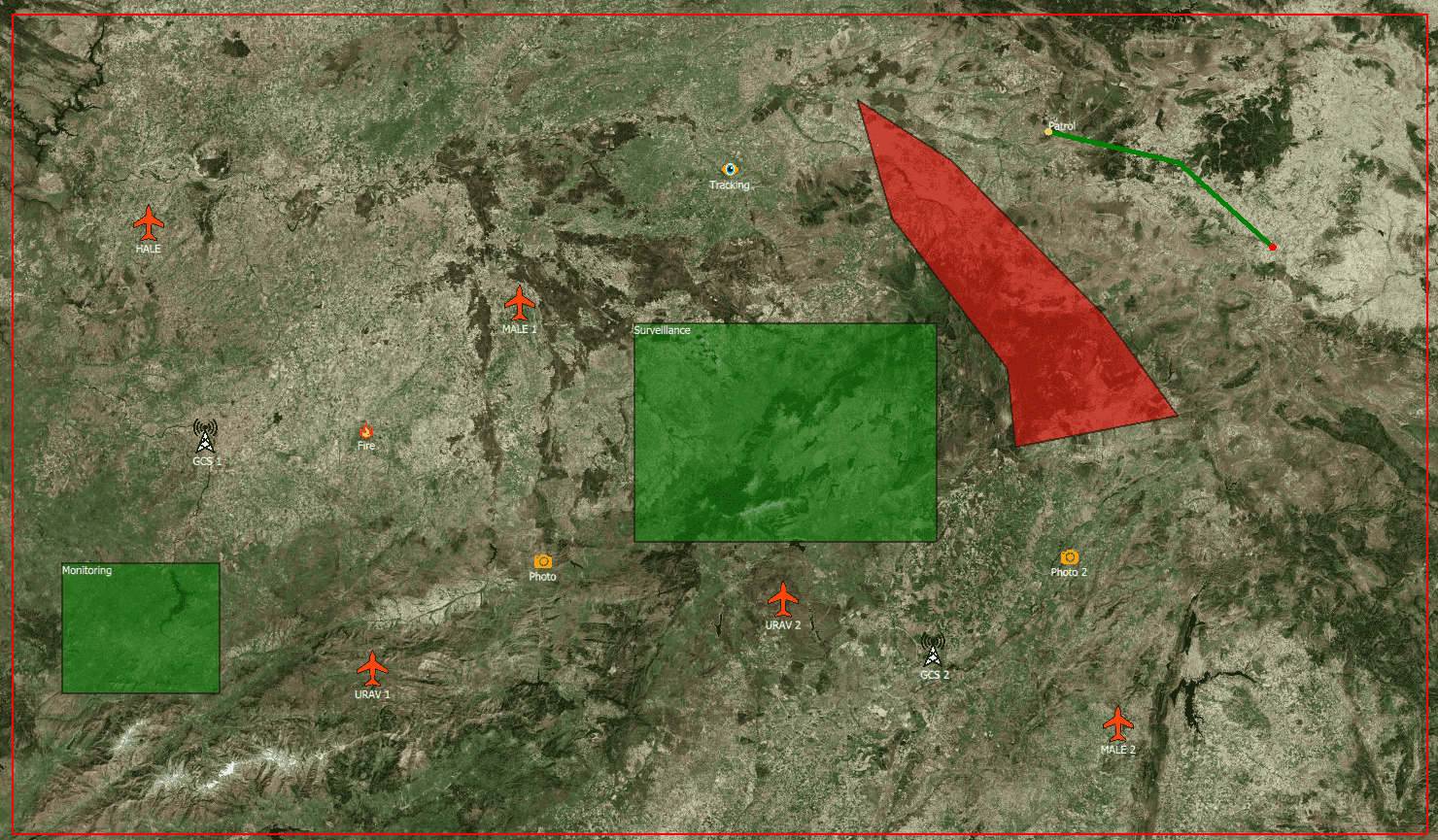}
        \caption{Mission 4.}
        \label{fig:d14}
    \end{subfigure}
    \quad
    \begin{subfigure}[b]{0.23\textwidth}
        \includegraphics[width=\textwidth]{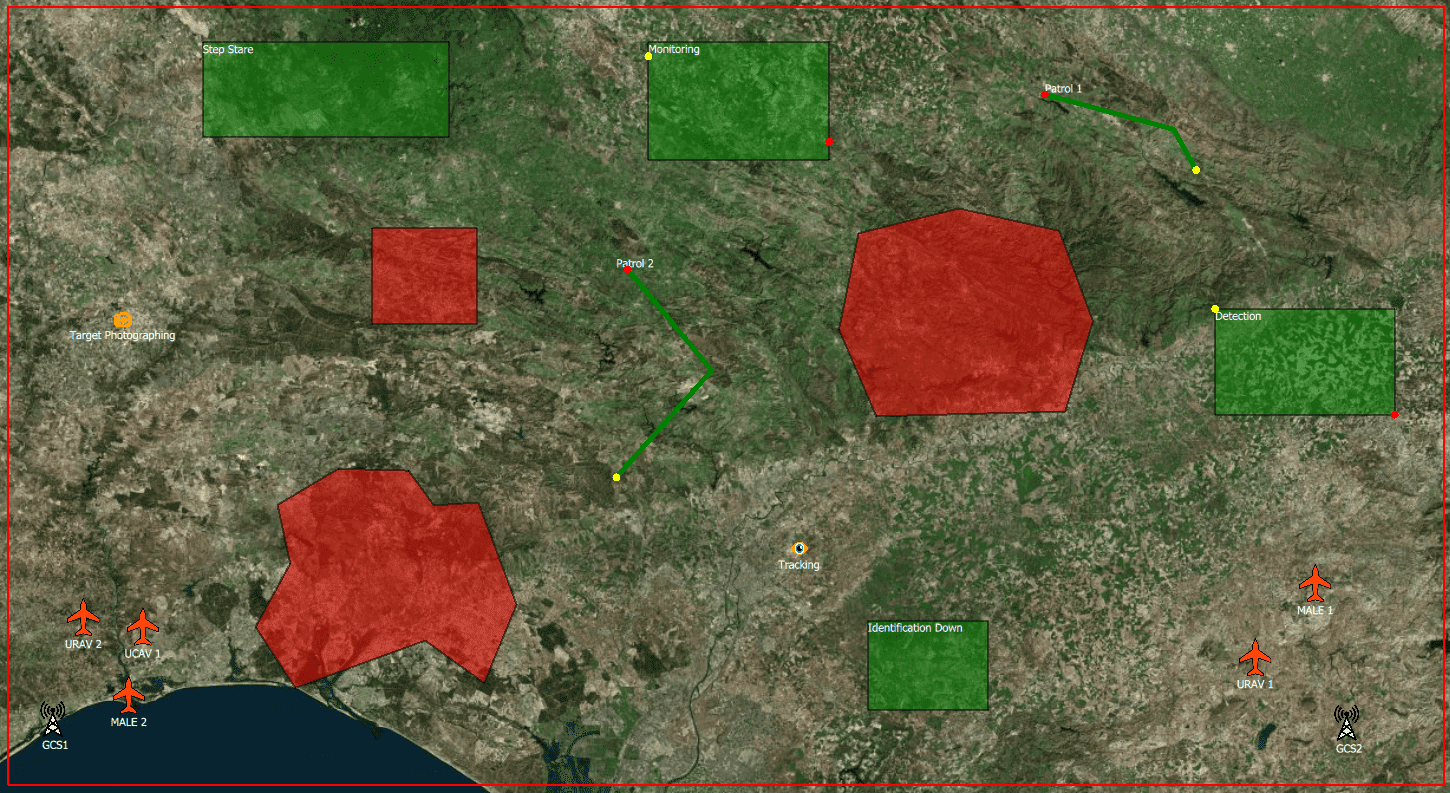}
        \caption{Mission 5.}
        \label{fig:d21}
    \end{subfigure}
    \quad
    \begin{subfigure}[b]{0.23\textwidth}
        \includegraphics[width=\textwidth]{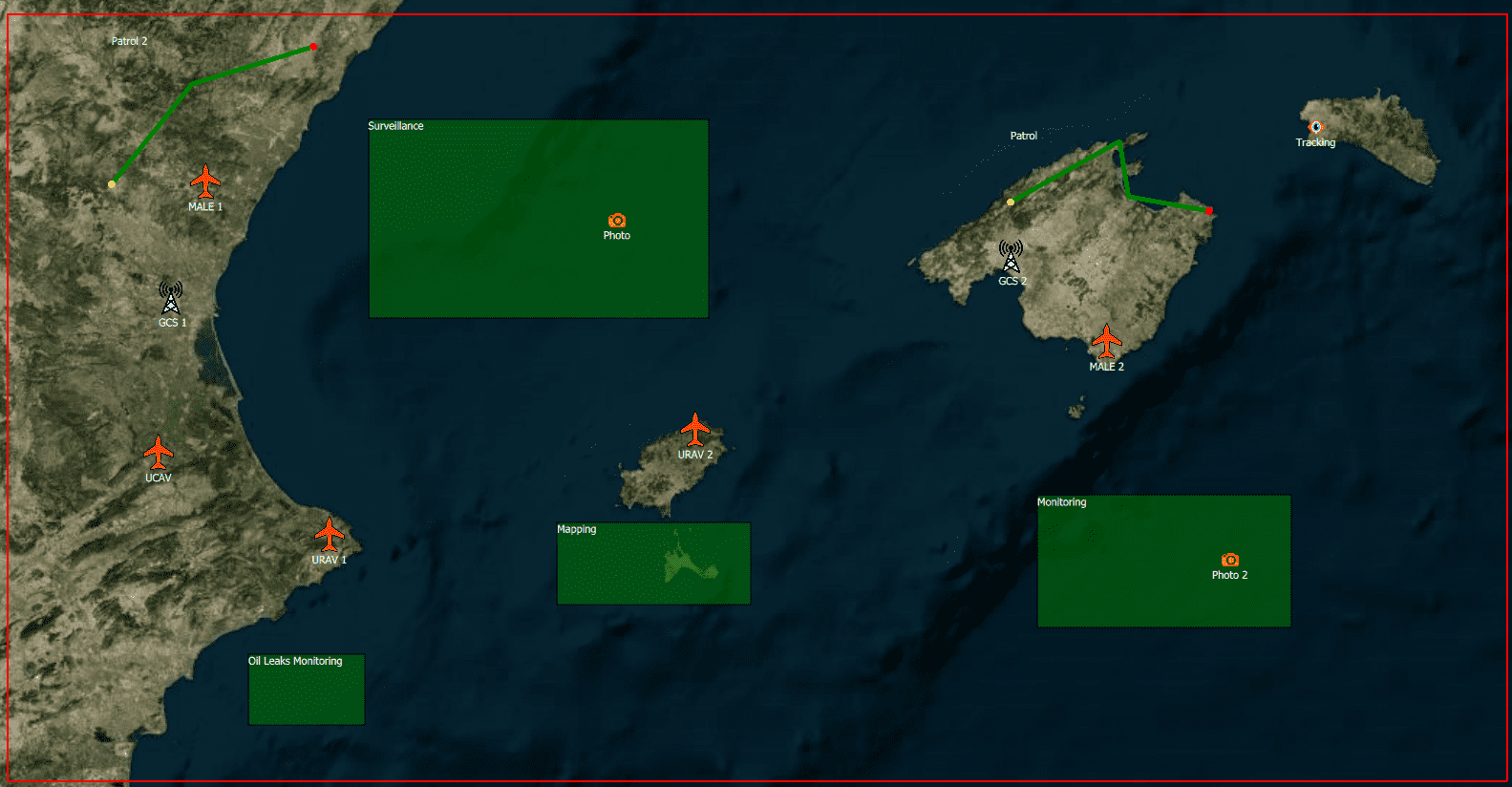}
        \caption{Mission 6.}
        \label{fig:d22}
    \end{subfigure}
    \quad
    \begin{subfigure}[b]{0.23\textwidth}
        \includegraphics[width=\textwidth]{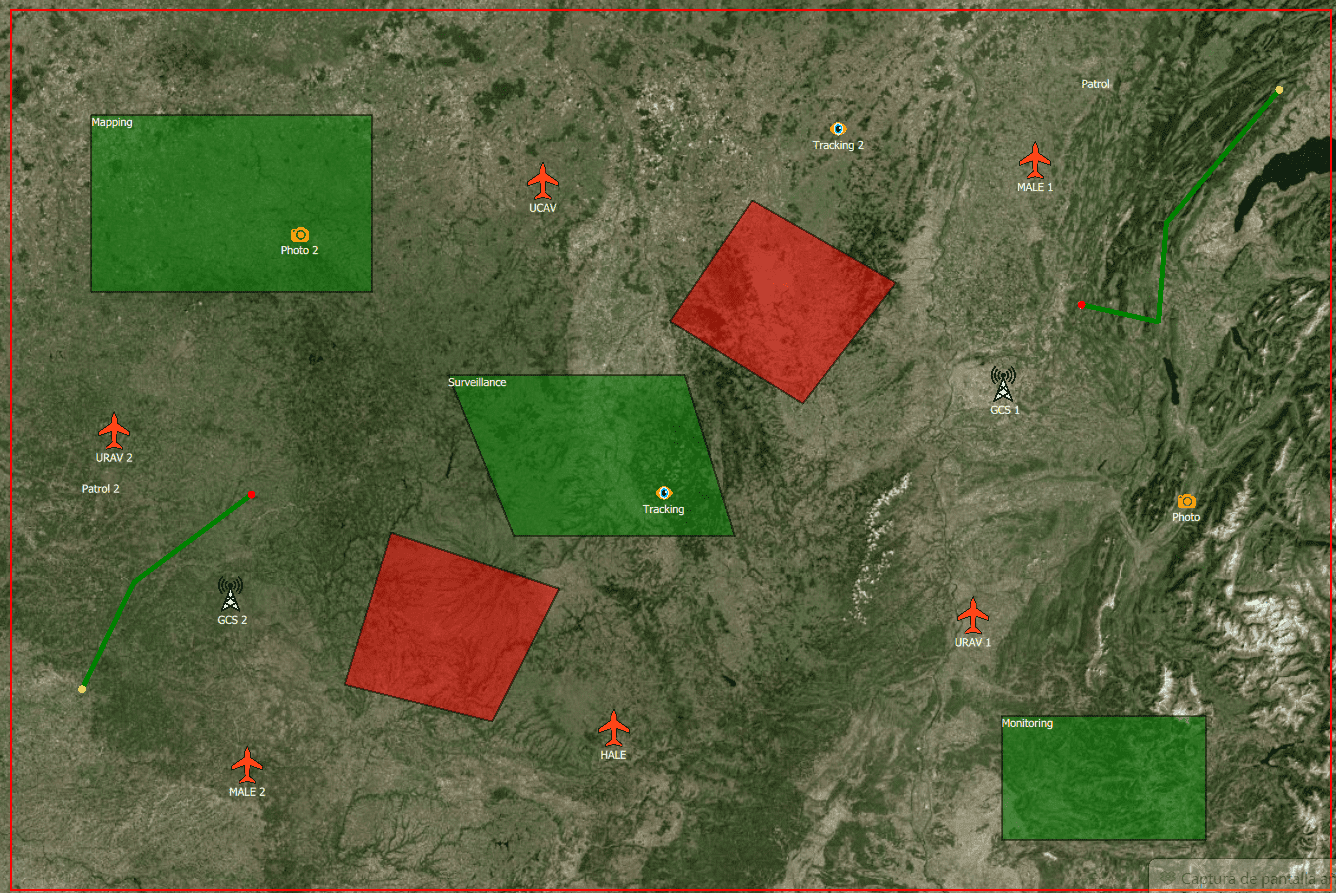}
        \caption{Mission 7.}
        \label{fig:d23}
    \end{subfigure}
    \quad
    \begin{subfigure}[b]{0.23\textwidth}
        \includegraphics[width=\textwidth]{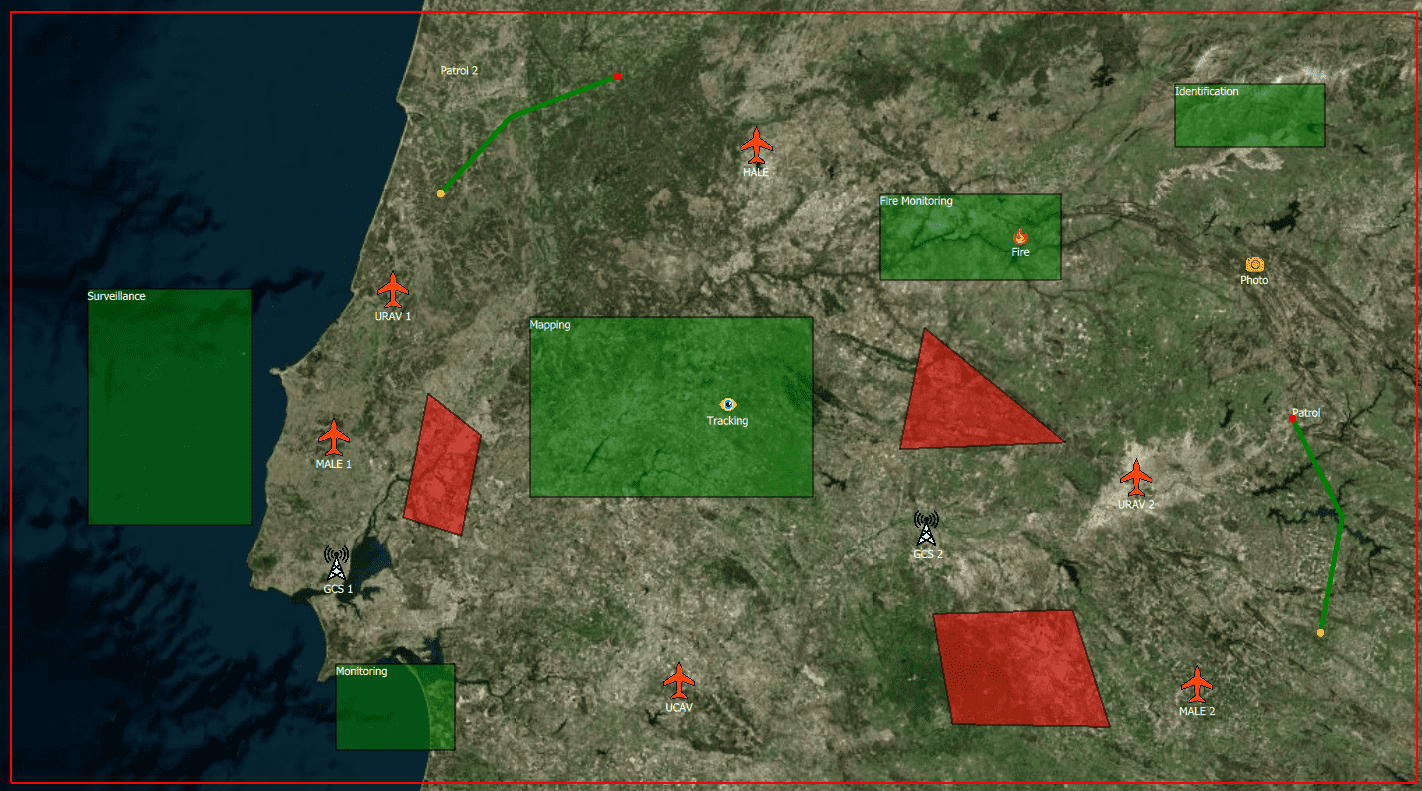}
        \caption{Mission 8.}
        \label{fig:d24}
    \end{subfigure}
    \quad
    \begin{subfigure}[b]{0.23\textwidth}
        \includegraphics[width=\textwidth]{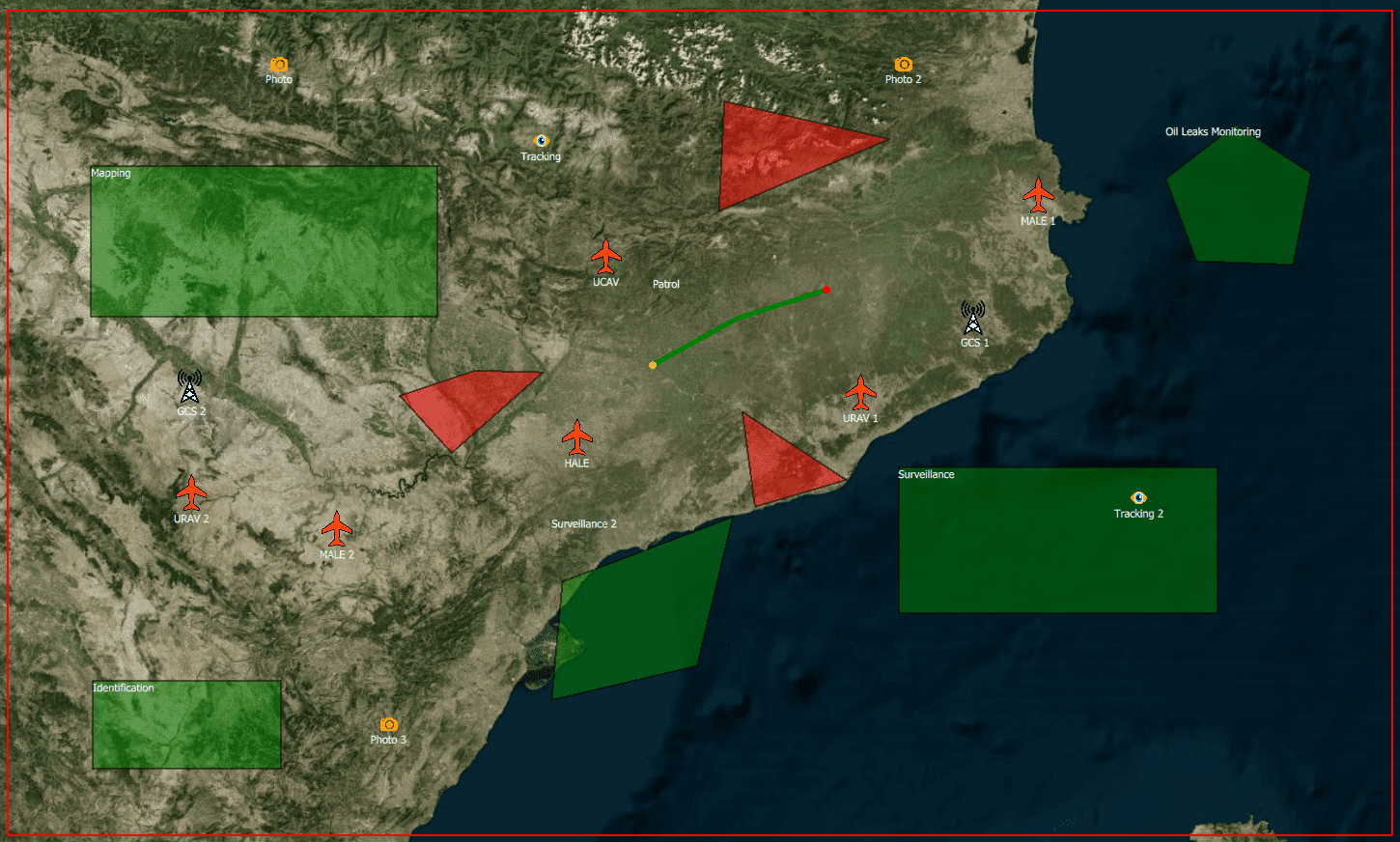}
        \caption{Mission 9.}
        \label{fig:d31}
    \end{subfigure}
    \quad
    \begin{subfigure}[b]{0.23\textwidth}
        \includegraphics[width=\textwidth]{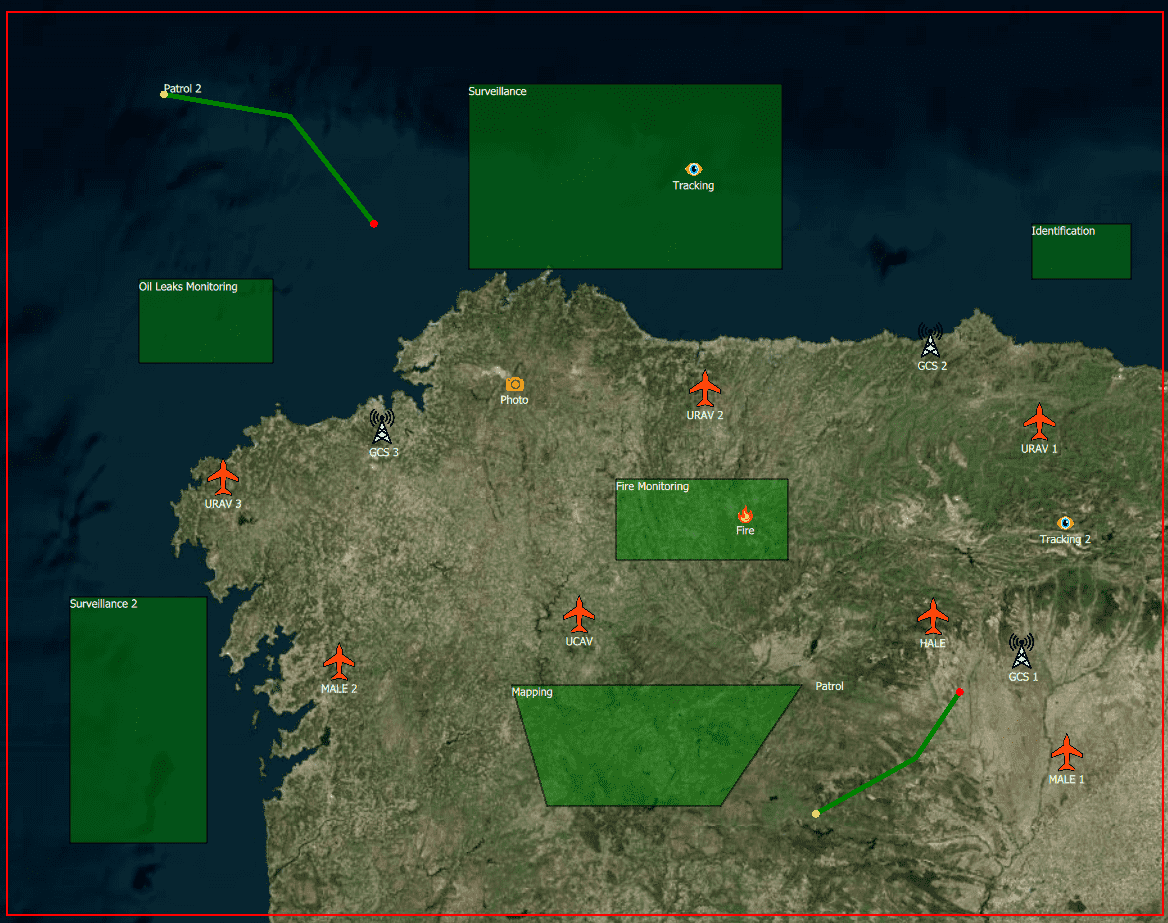}
        \caption{Mission 10.}
        \label{fig:d32}
    \end{subfigure}
    \quad
    \begin{subfigure}[b]{0.23\textwidth}
        \includegraphics[width=\textwidth]{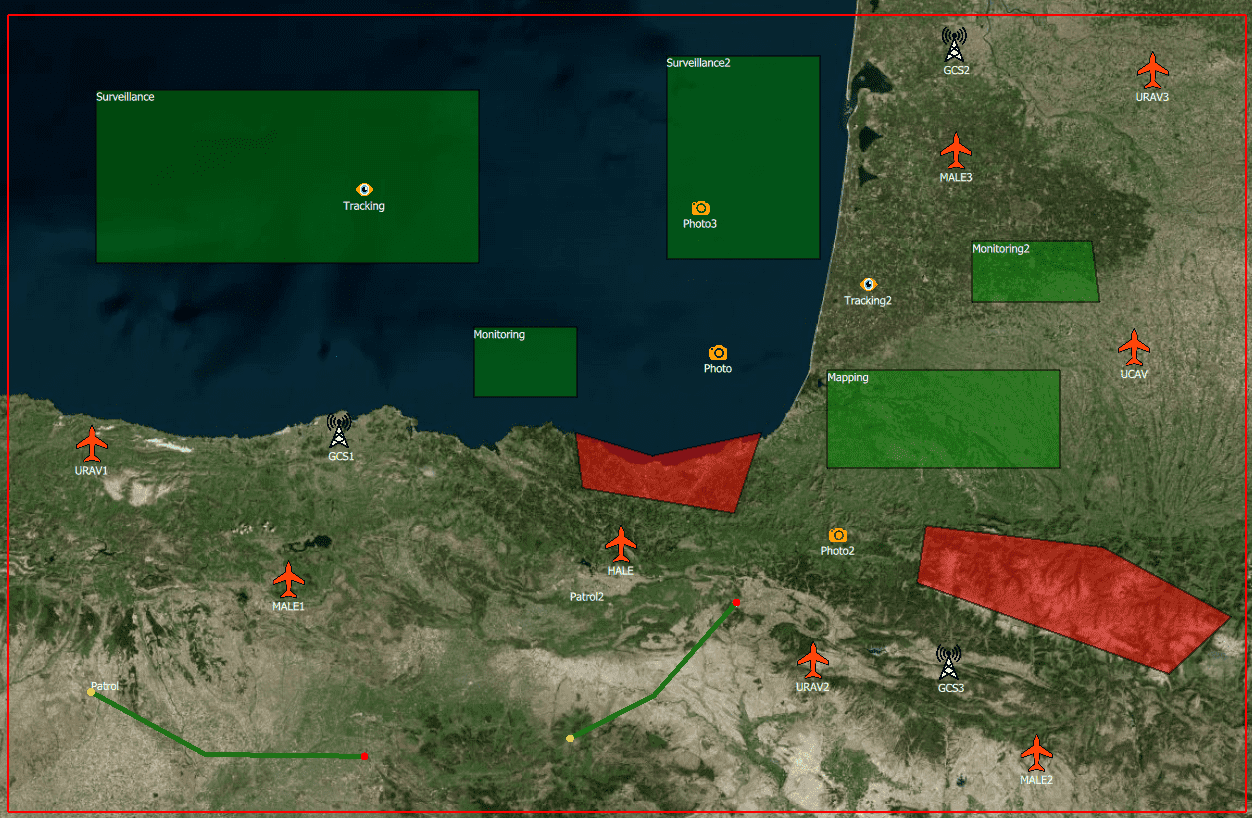}
        \caption{Mission 11.}
        \label{fig:d33}
    \end{subfigure}
    \quad
    \begin{subfigure}[b]{0.23\textwidth}
        \includegraphics[width=\textwidth]{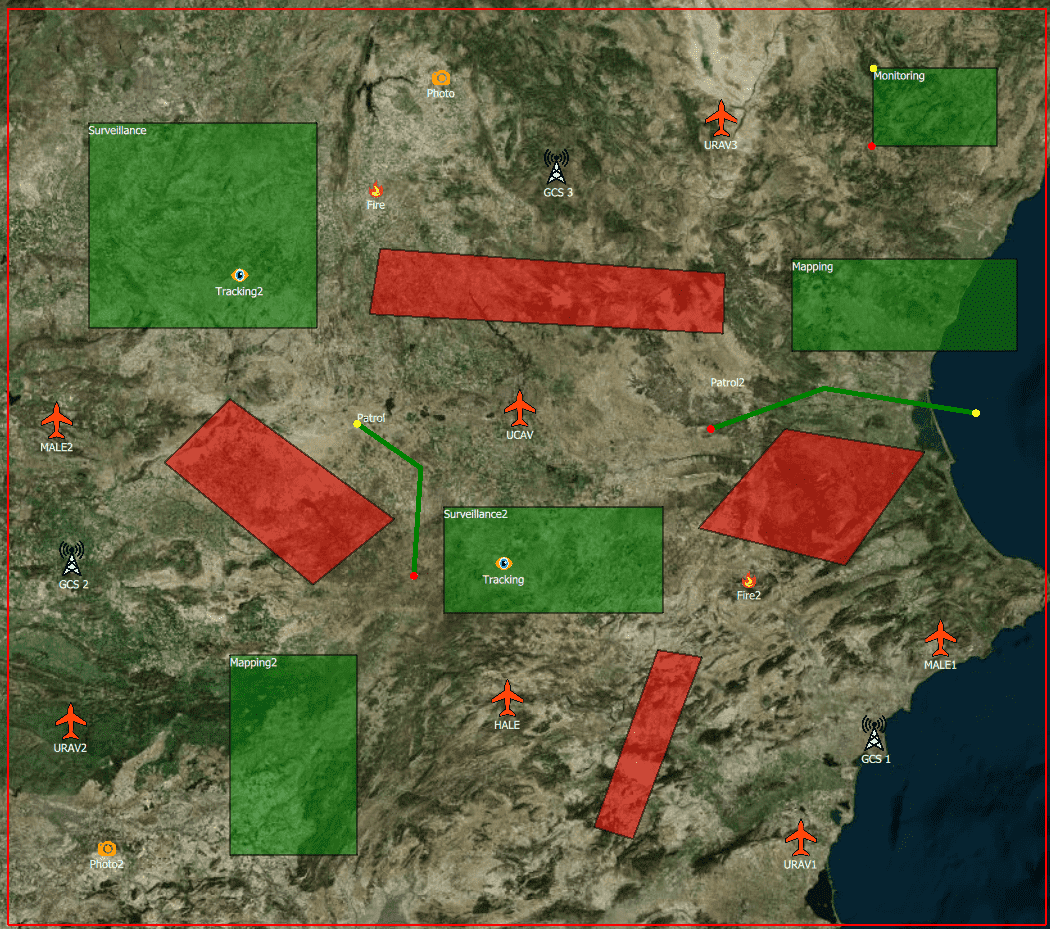}
        \caption{Mission 12.}
        \label{fig:d34}
    \end{subfigure}
    \caption{Mission Scenarios considered.}\label{fig:mission-scenario}
\end{figure*}

In these experiments, we tested the newly implemented \gls{skpnsga2} with 12 different scenarios, represented in Figure \ref{fig:mission-scenario}). In these figures, the green zones represent tasks, while the red zones represent \glspl{nfz}. There are also some point tasks represented with an icon, such as photographing, tracking or fire extinguishing. These scenarios are composed of an increasing number of tasks, multi-UAV tasks, \glspl{uav}, \glspl{gcs}, \glspl{nfz} and temporal dependencies between tasks (see Table \ref{tab:datasets}).

\begin{table}[!h]
\caption{Main features (number of UAVs, GCs, NFZs, etc.) for the different missions designed.}
\label{tab:datasets}
\begin{center}
\scalebox{0.72}{
\begin{tabular}{ccccccc }
  \toprule
  \begin{minipage}[t]{1cm}
  \centering
  Mission \\
  Id.
  \end{minipage} & Tasks & \begin{minipage}[t]{1.8cm}
  \centering
  Multi-UAV \\
  Tasks
  \end{minipage} & UAVs & GCSs & NFZs & \begin{minipage}[t]{2cm}
  \centering
  Time \\
  Dependencies
  \end{minipage}\\[3ex]
  \midrule
  1 & 5 & 0 & 3 & 1 & 0 & 0 \\
  2 & 6 & 1 & 3 & 1 & 1 & 0 \\
  3 & 6 & 1 & 4 & 2 & 2 & 1 \\
  4 & 7 & 1 & 5 & 2 & 1 & 2 \\
  5 & 8 & 2 & 5 & 2 & 3 & 1 \\
  6 & 9 & 2 & 5 & 2 & 0 & 2 \\
  7 & 9 & 2 & 6 & 2 & 2 & 2 \\
  8 & 10 & 2 & 6 & 2 & 3 & 3 \\
  9 & 11 & 3 & 6 & 2 & 3 & 2 \\
  10 & 12 & 3 & 7 & 3 & 0 & 2 \\
  11 & 12 & 3 & 8 & 3 & 2 & 3 \\
  12 & 13 & 4 & 7 & 3 & 4 & 4 \\
  \bottomrule
\end{tabular}
}
\end{center}
\end{table}

In the experiments, we tested these missions with the \gls{skpnsga2} algorithm developed in this work. In order to test the self-adaptation of the algorithm for the cone angle, the missions were also solved using \gls{nsga2} and the cone-angle-dependant implementation \gls{kpnsga2}\cite{Ramirez-Atencia2017} where the angle is fixed, using $120^{\circ}$, $135^{\circ}$, and $150^{\circ}$ angles (these approaches are names \gls{kpnsga2}-120, \gls{kpnsga2}-135 and \gls{kpnsga2}-150, respectively).

Each experiment has been executed 30 times, and the mean and standard deviation are presented for all the tables. On the other hand, the rest of parameters have been set to: population of the algorithm to 200, maximum number of generations to 300, the mutation probability to $5\%$, and the stopping criteria to 10.

\subsection{Experimental results}\label{experimentalresults}

To compare the results obtained, we computed the hypervolume with the normalized objectives for each solution set, taking as reference point the maximum point $(1,1,1,...)$. These results are shown in Table \ref{tab:results_hypervolume}. On the other hand, Table \ref{tab:results_num_solutions} shows the number of solutions obtained for each approach.

\begin{table}[ht]
\centering
\caption{Mean and standard deviation of the hypervolume obtained from the solutions given the different approaches for the different mission problems.} 
\label{tab:results_hypervolume}
\scalebox{0.72}{
\begin{tabular}{rlllll}
  \toprule
 Id. & NSGA-II & KPNSGA-II-120 & KPNSGA-II-135 & KPNSGA-II-150 & sKPNSGA-II \\ 
  \midrule
  1 & \cellcolor[HTML]{C0C0C0}$0.826\pm0.003$ & $0.79\pm0$ & $0.783\pm0.002$ & $0.782\pm0$ & $0.785\pm0.003$ \\ 
  2 & \cellcolor[HTML]{C0C0C0}$0.861\pm0.002$ & $0.72\pm0.036$ & $0.699\pm0.019$ & $0.693\pm0$ & $0.728\pm0.045$ \\ 
  3 & \cellcolor[HTML]{C0C0C0}$0.892\pm0.001$ & $0.79\pm0$ & $0.78\pm0.025$ & $0.78\pm0.025$ & $0.793\pm0.031$ \\ 
  4 & \cellcolor[HTML]{C0C0C0}$0.952\pm0.003$ & $0.889\pm0.063$ & $0.887\pm0.031$ & $0.886\pm0.029$ & $0.935\pm0.02$ \\ 
  5 & \cellcolor[HTML]{C0C0C0}$0.839\pm0.004$ & $0.741\pm0.062$ & $0.636\pm0.11$ & $0.605\pm0.091$ & $0.688\pm0.085$ \\ 
  6 & \cellcolor[HTML]{C0C0C0}$0.107\pm0.008$ & $0.095\pm0.002$ & $0.092\pm0.004$ & $0.088\pm0.004$ & $0.089\pm0.006$ \\ 
  7 & \cellcolor[HTML]{C0C0C0}$0.227\pm0.012$ & $0.209\pm0.006$ & $0.203\pm0.012$ & $0.195\pm0.018$ & $0.201\pm0.014$ \\ 
  8 & \cellcolor[HTML]{C0C0C0}$0.161\pm0.015$ & $0.14\pm0.005$ & $0.136\pm0.004$ & $0.132\pm0.006$ & $0.138\pm0.004$ \\ 
  9 & \cellcolor[HTML]{C0C0C0}$0.122\pm0.013$ & $0.095\pm0.008$ & $0.091\pm0.007$ & $0.087\pm0.007$ & $0.091\pm0.003$ \\ 
  10 & \cellcolor[HTML]{C0C0C0}$0.147\pm0.017$ & $0.123\pm0.017$ & $0.104\pm0.015$ & $0.107\pm0.013$ & $0.114\pm0.017$ \\ 
  11 & \cellcolor[HTML]{C0C0C0}$0.161\pm0.020$ & $0.138\pm0.011$ & $0.124\pm0.008$ & $0.117\pm0.01$ & $0.134\pm0.014$ \\ 
  12 & \cellcolor[HTML]{C0C0C0}$0.151\pm0.022$ & $0.132\pm0.011$ & $0.118\pm0.016$ & $0.11\pm0.016$ & $0.123\pm0.008$ \\
   \bottomrule
\end{tabular}
}
\end{table}

\begin{table}[ht]
\centering
\caption{Mean and standard deviation of the number of solutions obtained from the different approaches for the different mission problems.} 
\label{tab:results_num_solutions}
\scalebox{0.72}{
\begin{tabular}{rlllll}
  \toprule
 Id. & NSGA-II & KPNSGA-II-120 & KPNSGA-II-135 & KPNSGA-II-150 & sKPNSGA-II \\ 
  \midrule
1 & $89.32\pm5.46$ & $2\pm0$ & $1.05\pm0.22$ & \cellcolor[HTML]{C0C0C0}$1\pm0$ & $1.3\pm0.47$ \\ 
  2 & $432.13\pm10.51$ & $1.44\pm0.51$ & $1.13\pm0.35$ & \cellcolor[HTML]{C0C0C0}$1.06\pm0.25$ & $2.71\pm2.52$ \\ 
  3 & $517.53\pm29.92$ & \cellcolor[HTML]{C0C0C0}$1\pm0$ & \cellcolor[HTML]{C0C0C0}$1\pm0$ & \cellcolor[HTML]{C0C0C0}$1\pm0$ & $1.36\pm0.95$ \\ 
  4 & $322.61\pm42.7$ & $2.7\pm0.92$ & $1.55\pm0.51$ & $1.5\pm0.51$ & \cellcolor[HTML]{C0C0C0}$1.06\pm0.24$ \\ 
  5 & $342.24\pm55.43$ & $3.3\pm1.56$ & $1.9\pm0.99$ & \cellcolor[HTML]{C0C0C0}$1.5\pm0.83$ & $2.65\pm1.66$ \\ 
  6 & $792.21\pm83.12$ & $4.04\pm0.81$ & $2.24\pm0.72$ & \cellcolor[HTML]{C0C0C0}$1.6\pm0.76$ & $2.43\pm2.92$ \\ 
  7 & $1191.25\pm116.74$ & $4.08\pm1.79$ & $2.84\pm1.21$ & \cellcolor[HTML]{C0C0C0}$2.08\pm0.76$ & $3.67\pm3.94$ \\ 
  8 & $692.98\pm78.85$ & $3.88\pm3.89$ & $1.8\pm1.19$ & \cellcolor[HTML]{C0C0C0}$1.24\pm0.52$ & $2.07\pm1.27$ \\ 
  9 & $822.94\pm115.24$ & $7.58\pm3.6$ & $3.72\pm2.82$ & \cellcolor[HTML]{C0C0C0}$1.8\pm1.16$ & $4.75\pm1.5$ \\ 
  10 & $579.25\pm98.83$ & $11.75\pm7.71$ & $3.84\pm3.04$ & \cellcolor[HTML]{C0C0C0}$3.12\pm1.81$ & $4.5\pm4.95$ \\ 
  11 & $967.65\pm126.79$ & $17.83\pm13.66$ & $8.4\pm5.8$ & \cellcolor[HTML]{C0C0C0}$3.8\pm1.56$ & $8.75\pm9.07$ \\ 
  12 & $484.52\pm76.6$ & $6.33\pm3.84$ & $3.52\pm2.45$ & $1.84\pm1.21$ & \cellcolor[HTML]{C0C0C0}$1.36\pm0.5$ \\ 
   \bottomrule
\end{tabular}
}
\end{table}

In these results, it is appreciable how the hypervolume decreases with bigger angles, as well as the number of solutions. On the other hand, it is appreciable how \gls{nsga2} gets worse results as the complexity of the problem grows (the difference of hypervolume with respect to \gls{skpnsga2} decreases), due to the big number of solutions composing the \gls{pof}.

In order to measure these hypervolume and number of solutions together, the HDist metric (see Section \ref{hypervolumedistribution}) is used. The values of this metric for each result are presented in Table \ref{tab:results_hdist}. With this, we can clearly appreciate that \gls{skpnsga2} gets the best results for this metric, as it has been optimized during the evolutionary process. In addition, we have computed the Wilcoxon test \cite{Hollander2014}, comparing \gls{skpnsga2} with the rest of approaches. The test succeed in all problems, with a $p-value < 0.05$.

\begin{table}[ht]
\centering
\caption{Mean and standard deviation of the HDist metric obtained from the solutions of the different approaches for the different mission problems.}
\label{tab:results_hdist}
\scalebox{0.72}{
\begin{tabular}{rlllll}
  \toprule
 Id. & NSGA-II & KPNSGA-II-120 & KPNSGA-II-135 & KPNSGA-II-150 & sKPNSGA-II \\ 
  \midrule
1 & $0.046\pm0.059$ & \cellcolor[HTML]{C0C0C0}$0.631\pm0$ & $0.587\pm0.01$ & $0.585\pm0$ & $0.599\pm0.022$ \\ 
  2 & $0.02\pm0.024$ & $0.16\pm0.213$ & $0.035\pm0.113$ & $0\pm0$ & \cellcolor[HTML]{C0C0C0}$0.207\pm0.261$ \\ 
  3 & $0.095\pm0.052$ & $0.397\pm0$ & $0.337\pm0.145$ & $0.337\pm0.145$ & \cellcolor[HTML]{C0C0C0}$0.413\pm0.181$ \\ 
  4 & $0.201\pm0.103$ & $0.777\pm0.205$ & $0.773\pm0.098$ & $0.771\pm0.094$ & \cellcolor[HTML]{C0C0C0}$0.93\pm0.065$ \\ 
  5 & $0.194\pm0.126$ & \cellcolor[HTML]{C0C0C0}$0.702\pm0.161$ & $0.428\pm0.288$ & $0.346\pm0.241$ & $0.564\pm0.223$ \\ 
  6 & $0.102\pm0.098$ & \cellcolor[HTML]{C0C0C0}$0.801\pm0.035$ & $0.737\pm0.067$ & $0.678\pm0.076$ & $0.692\pm0.095$ \\ 
  7 & $0.088\pm0.106$ & \cellcolor[HTML]{C0C0C0}$0.807\pm0.064$ & $0.735\pm0.135$ & $0.653\pm0.197$ & $0.711\pm0.156$ \\ 
  8 & $0.064\pm0.056$ & \cellcolor[HTML]{C0C0C0}$0.534\pm0.115$ & $0.447\pm0.092$ & $0.355\pm0.137$ & $0.492\pm0.085$ \\ 
  9 & $0.078\pm0.043$ & \cellcolor[HTML]{C0C0C0}$0.387\pm0.146$ & $0.299\pm0.134$ & $0.233\pm0.135$ & $0.31\pm0.053$ \\ 
  10 & $0.054\pm0.026$ & \cellcolor[HTML]{C0C0C0}$0.701\pm0.183$ & $0.499\pm0.17$ & $0.535\pm0.156$ & $0.607\pm0.198$ \\ 
  11 & $0.021\pm0.013$ & \cellcolor[HTML]{C0C0C0}$0.714\pm0.131$ & $0.552\pm0.101$ & $0.469\pm0.118$ & $0.668\pm0.17$ \\ 
  12 & $0.009\pm0.012$ & \cellcolor[HTML]{C0C0C0}$0.818\pm0.1$ & $0.692\pm0.15$ & $0.617\pm0.144$ & $0.736\pm0.069$ \\ 
   \bottomrule
\end{tabular}
}
\end{table}

\begin{figure*}[!h]
		\includegraphics[width=\textwidth]{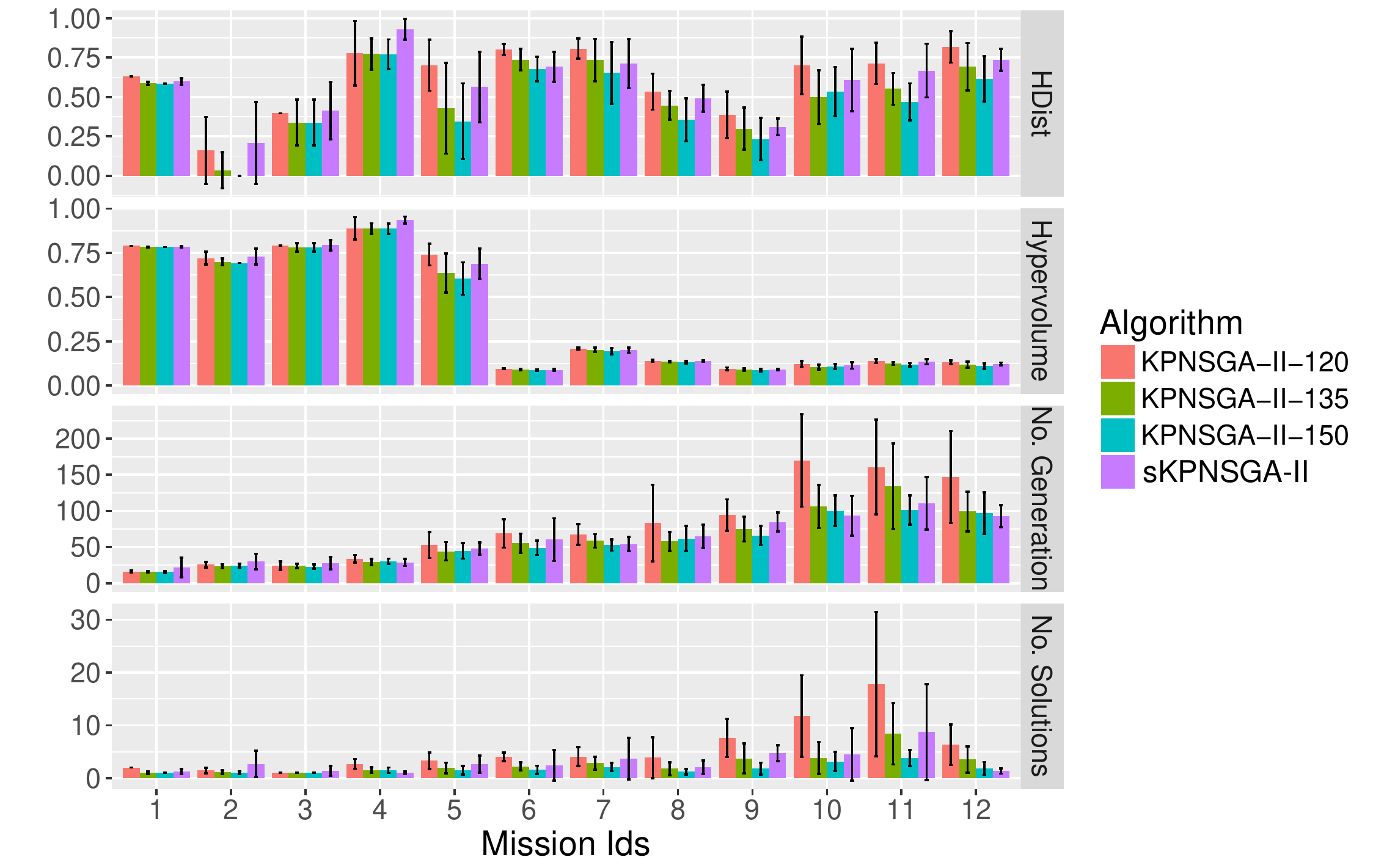}
		\centering
		\caption{Comparison of hypervolume, number of solutions obtained, HDist metric and number of generations needed to converge for the different approaches in each problem.}
		\label{fig:comparisonMetrics}
	\end{figure*}

The results also show that the HDist metric presents a bigger standard deviation in \gls{skpnsga2} than in the rest of approaches. This can be better seen in the HDist graphic in Figure \ref{fig:comparisonMetrics}. This is specially appreciable in the most complex problems, and it is due to the early start of the golden section search algorithm due to the condition of the high number of solutions (see Algorithm \ref{alg:golden-section}, Line 7). Erasing this condition, will outperform the convergence of the approach, but at the expense of increasing the number of generations needed to converge and, consequently, the runtime of the algorithm.

Table \ref{tab:results_num_generations} shows the number of generations needed to converge for the different missions and algorithms. Here, it is shown that the runtime of the algorithm is also reduced with \gls{skpnsga2} compared to \gls{nsga2}, which in most cases was not even able to converge in the maximum number of generations defined. On the other hand, the runtime of \gls{skpnsga2} is bigger than the approaches where the angle is fixed. Concretely, the higher the cone angle, the faster the algorithm.

\begin{table}[ht]
\centering
\caption{Mean and standard deviation of the number of generations needed to converge in the different approaches for the different mission problems.} 
\label{tab:results_num_generations}
\scalebox{0.65}{
\begin{tabular}{rlllll}
  \toprule
 Id. & NSGA-II & KPNSGA-II-120 & KPNSGA-II-135 & KPNSGA-II-150 & sKPNSGA-II \\ 
  \midrule
  1 & $64.789\pm14.273$ & $16.048\pm1.244$ & $15.762\pm1.091$ & \cellcolor[HTML]{C0C0C0}$15.571\pm1.207$ & $21.75\pm13.78$ \\ 
  2 & $271.875\pm49.295$ & $25.688\pm3.737$ & \cellcolor[HTML]{C0C0C0}$23.267\pm2.712$ & $24.188\pm2.664$ & $29.941\pm10.802$ \\ 
  3 & $298.124\pm2.54$ & $24.25\pm6.008$ & $23.9\pm2.936$ & \cellcolor[HTML]{C0C0C0}$23\pm2.974$ & $27.909\pm8.717$ \\ 
  4 & $300\pm0$ & $33.75\pm5.466$ & $29.15\pm4.38$ & $30.6\pm3.47$ & \cellcolor[HTML]{C0C0C0}$28.722\pm4.812$ \\ 
  5 & $300\pm0$ & $53\pm18.061$ & \cellcolor[HTML]{C0C0C0}$44.1\pm12.732$ & $44.8\pm10.665$ & $48.176\pm8.465$ \\ 
  6 & $300\pm0$ & $68.917\pm19.64$ & $55.4\pm13.292$ & \cellcolor[HTML]{C0C0C0}$49.12\pm10.035$ & $60.286\pm29.508$ \\ 
  7 & $300\pm0$ & $67.5\pm14.347$ & $58.6\pm9.341$ & \cellcolor[HTML]{C0C0C0}$52.84\pm7.867$ & $54.111\pm9.752$ \\ 
  8 & $300\pm0$ & $83.125\pm53.161$ & \cellcolor[HTML]{C0C0C0}$57.84\pm13.243$ & $61.88\pm17.302$ & $65.071\pm16.074$ \\ 
  9 & $300\pm0$ & $94.167\pm21.908$ & $74.96\pm17.155$ & \cellcolor[HTML]{C0C0C0}$66\pm13.279$ & $84.5\pm13.077$ \\ 
  10 & $300\pm0$ & $169.958\pm63.995$ & $106.12\pm29.914$ & $100.44\pm21.389$ & \cellcolor[HTML]{C0C0C0}$93.5\pm27.577$ \\ 
  11 & $300\pm0$ & $160.792\pm65.543$ & $134.48\pm59.178$ & \cellcolor[HTML]{C0C0C0}$101.2\pm20.114$ & $110.5\pm36.189$ \\ 
  12 & $300\pm0$ & $146.917\pm63.827$ & $99.32\pm27.436$ & $97.12\pm28.642$ & \cellcolor[HTML]{C0C0C0}$92.714\pm15.339$ \\ 
   \bottomrule
\end{tabular}
}
\end{table}

\begin{figure*}[!t]
    \centering
    \begin{subfigure}[b]{0.47\textwidth}
        \includegraphics[width=\textwidth]{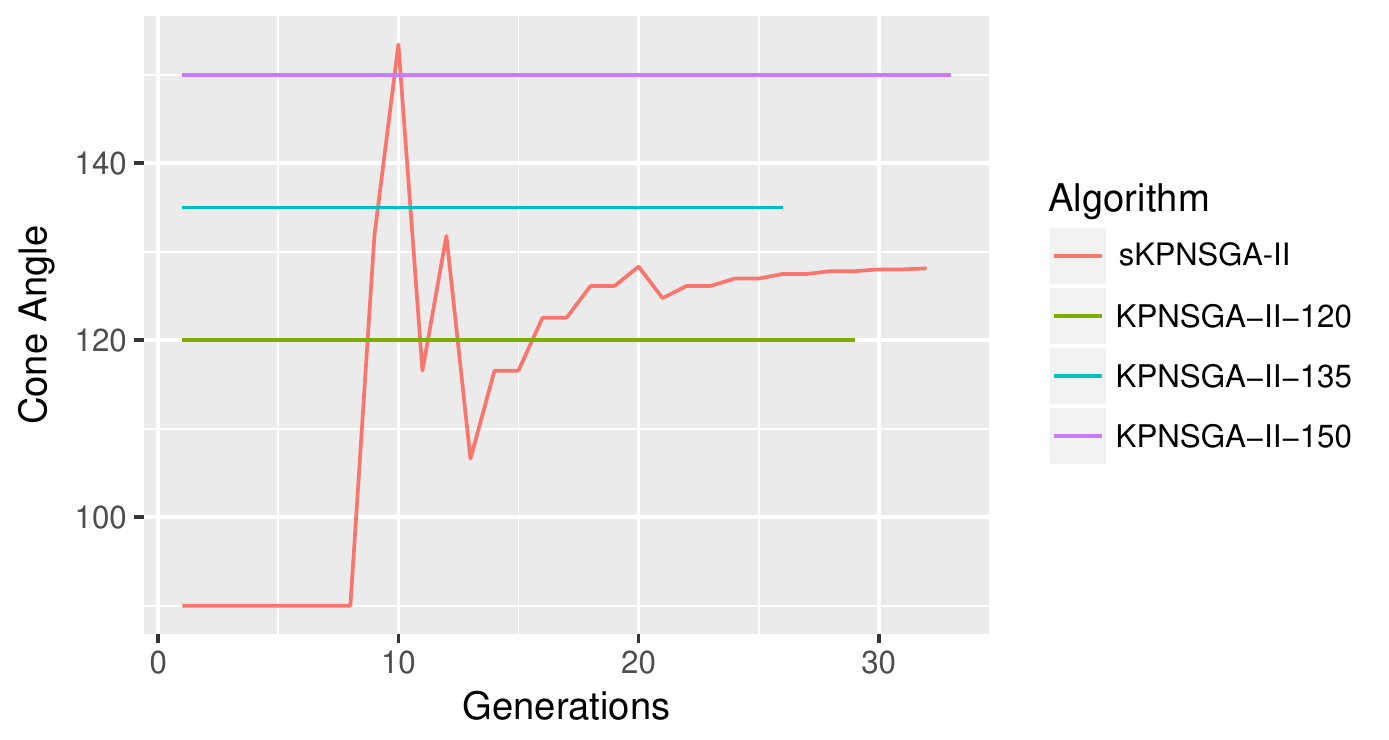}
        \caption{Cone angle.}
        \label{fig:cone-angle-gen}
    \end{subfigure}
    \quad
    \begin{subfigure}[b]{0.47\textwidth}
        \includegraphics[width=\textwidth]{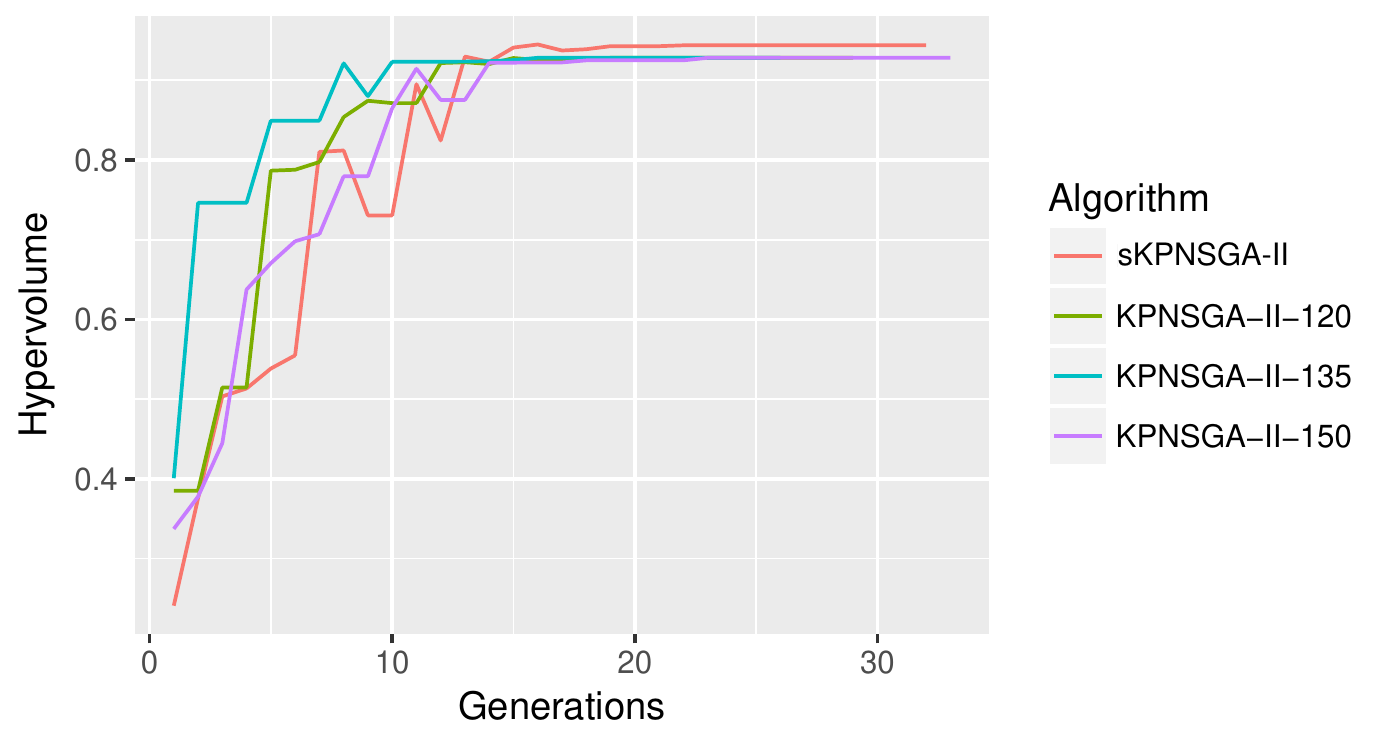}
        \caption{Hypervolume.}
        \label{fig:hyp-gen}
    \end{subfigure}
    \quad
    \begin{subfigure}[b]{0.47\textwidth}
        \includegraphics[width=\textwidth]{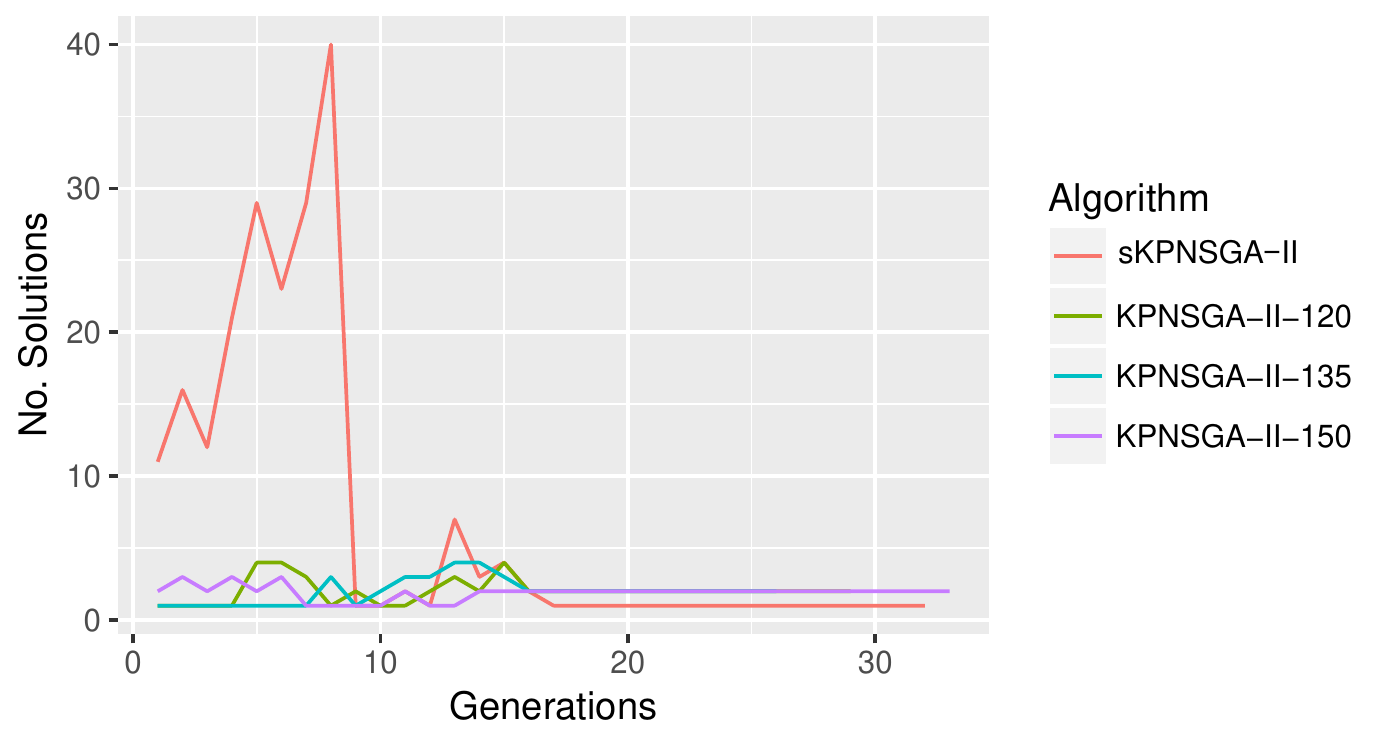}
        \caption{Number of solutions.}
        \label{fig:numsols-gen}
    \end{subfigure}
    \quad
    \begin{subfigure}[b]{0.47\textwidth}
        \includegraphics[width=\textwidth]{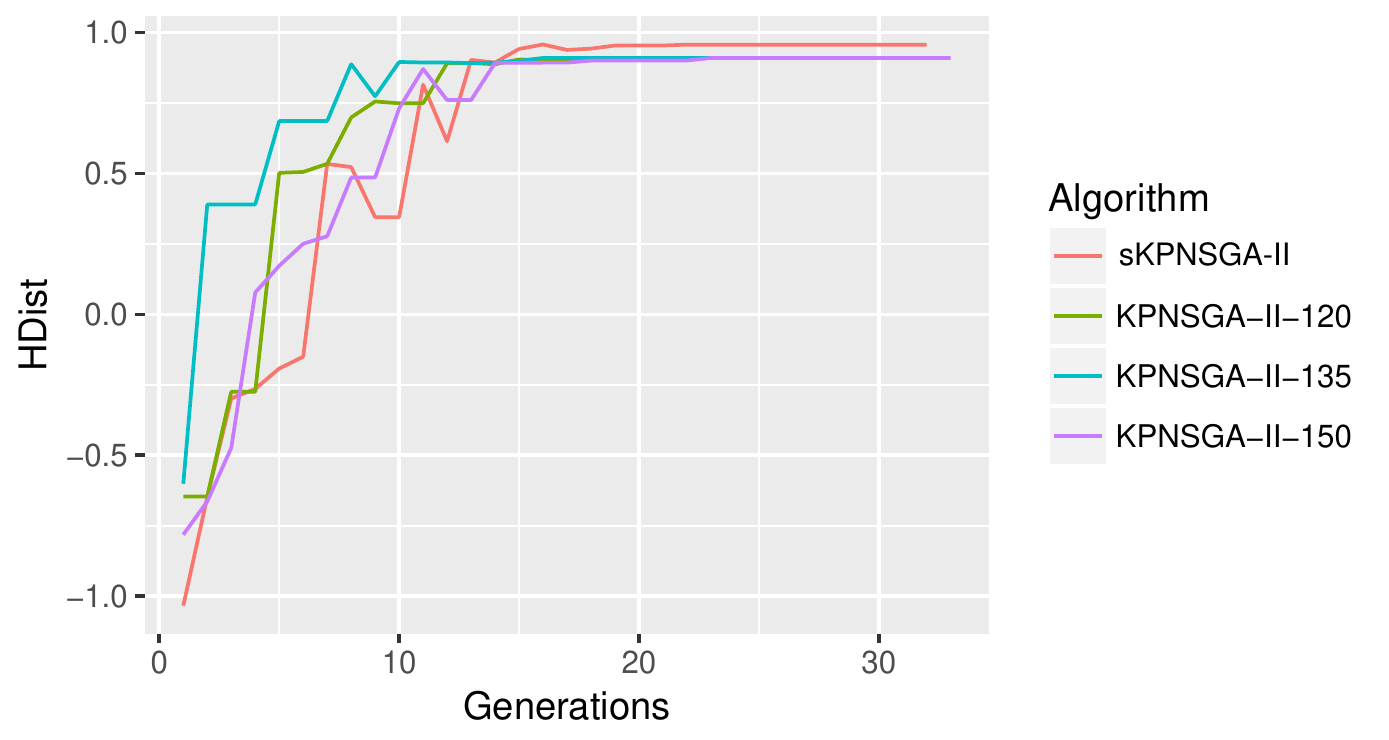}
        \caption{HDist metric.}
        \label{fig:hdist-gen}
    \end{subfigure}
    \caption{Evolution of metrics by generation for mission 4.}\label{fig:results-gen}
\end{figure*}

In order to observe how \gls{skpnsga2} evolves, Figure \ref{fig:results-gen} shows the evolution of Cone Angle, Hypervolume, Number of solutions obtained and HDist Metric by generation in Mission 4, and compares them with the fix angle approaches. Here, it is appreciable how the cone angle critically varies when the golden section search starts but rapidly converge to the optimum value. In the HDist graphic, it is shown how \gls{skpnsga2} starts with worse HDist than the other approaches as it has not determined its cone angle yet, but once it does, it get the better result.

\begin{figure*}[!t]
    \centering
    \begin{subfigure}[b]{0.47\textwidth}
        \includegraphics[width=\textwidth]{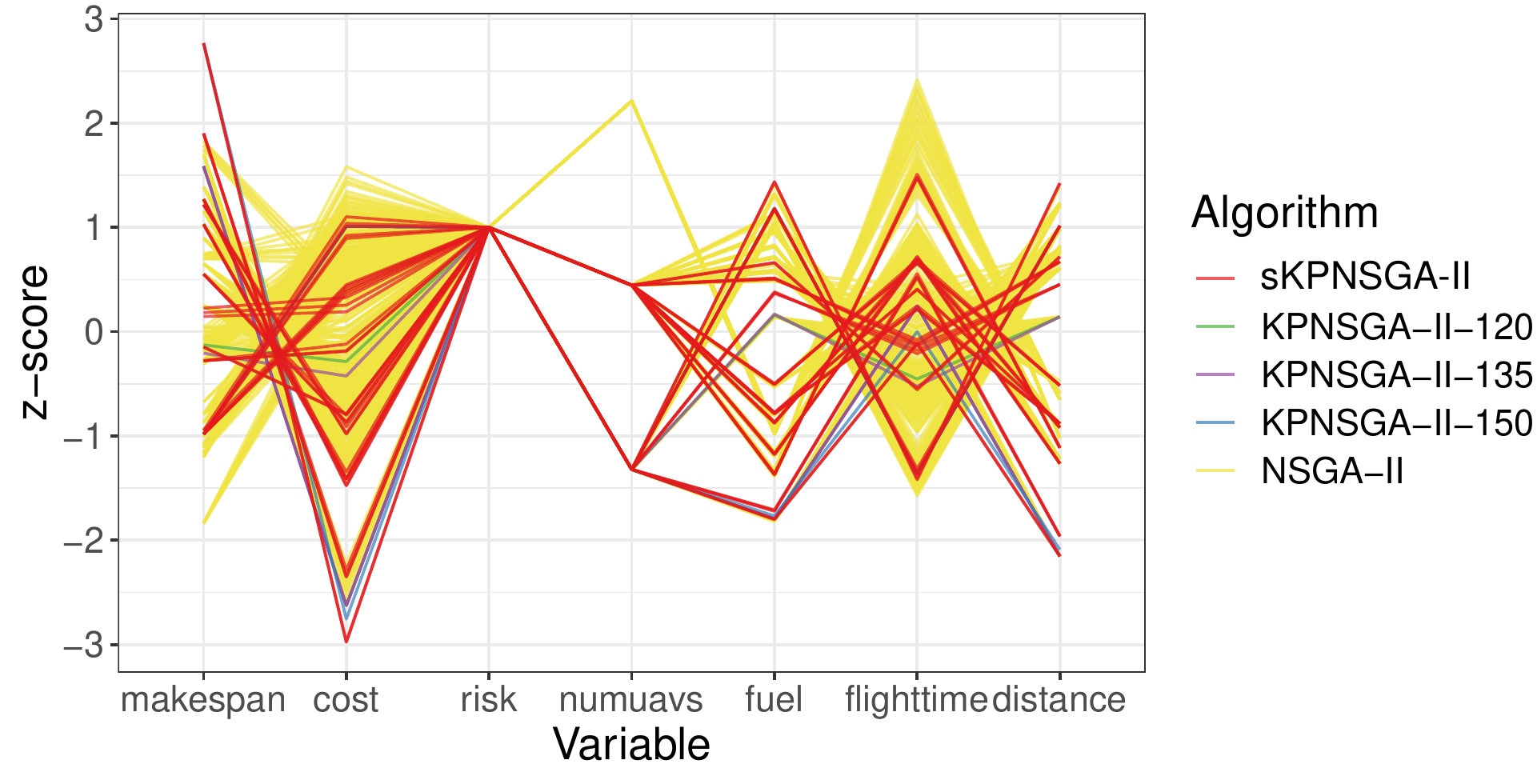}
        \caption{Parallel plot.}
        \label{fig:parallel-plot}
    \end{subfigure}
    \quad
    \begin{subfigure}[b]{0.47\textwidth}
        \includegraphics[width=\textwidth]{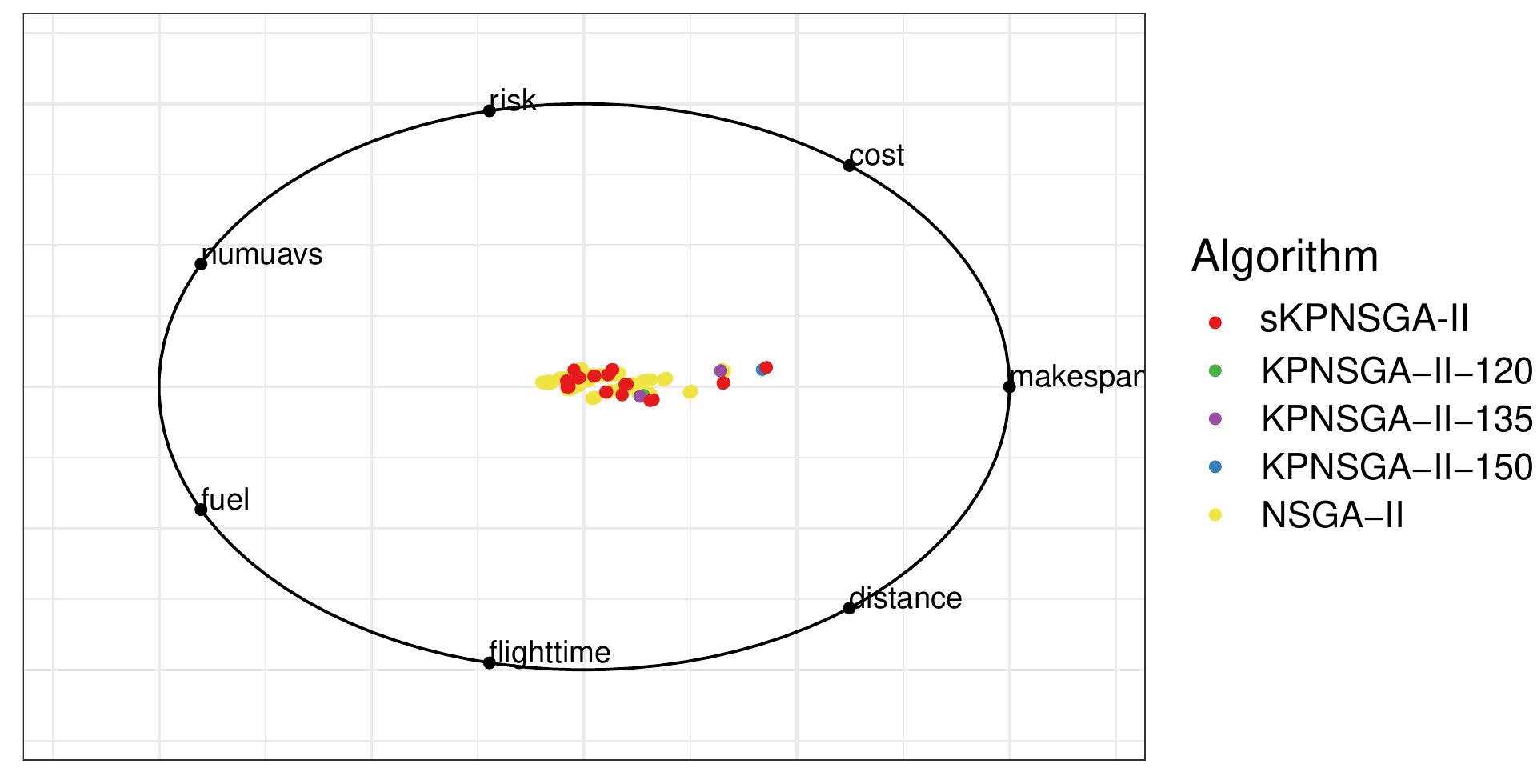}
        \caption{Radial plot.}
        \label{fig:rad-plot}
    \end{subfigure}
    \caption{Visualization of solutions for mission 4.}\label{fig:solutions-plot}
\end{figure*}

On the other hand, Figure \ref{fig:solutions-plot} shows the parallel and radial plot of the solutions obtained by each approach. Here, it can be seen how the solutions obtained by \gls{skpnsga2} are spread with all the optimization variables, proving that the solutions obtained are a significant sample of the solutions of the \gls{pof}.

\section{Conclusion}
In this work, we have presented an extension of the \gls{nsga2} algorithm based on Knee Points in order to guide and focus the search process of the algorithm for significant solutions. To do that, we have presented the concept of cone-domination, which substitutes the domination concept in the algorithm. The algorithm uses a cone angle which self-adapts in the algorithm using the golden section search.

This new approach have been tested with real Multi-UAV Mission Planning Problems, which are complex and have a lot of solutions. In these problems, the mission operator has to select the best solution among all the obtained, so reducing the number of solutions and present just the most significant ones to the operator will reduce its workload.

In the experimental phase, the approach has been compared against \gls{nsga2} and non-self-adapting approaches with three different cone angles (120, 135 and 150). The results showed that the \gls{skpnsga2} approach adapts the angle according to the HDist metric, which is clearly maximized when compared to the other approaches. The number of solutions returned are quite small while the most of the hypervolume is maintained compared to \gls{nsga2}.

On the other hand, the results obtained from the experimental phase showed that the number of generations needed to converge are also improved by the new algorithm compared to \gls{nsga2}, while the fixed angle approaches converge earlier. In the most complex problems, \gls{nsga2} could not find the complete \gls{pof}, while \gls{skpnsga2} could converge.

In our future research works, and in order to improve the decision making process for the operator, we will also develop some ranker algorithm for the solutions returned by \gls{skpnsga2}, which allows to easily select the most interesting and relevant solutions to human operators.


%



\section*{Acknowledgment}
This work has been supported by Airbus Defence \& Space (under grants number: FUAM-076914 and FUAM-076915), and by the next research projects: DeepBio (TIN2017-85727-C4-3-P), funded by Spanish Ministry of Economy and Competitivity (MINECO), and CYNAMON (CAM grant S2018/TCS-4566), under the European Regional Development Fund FEDER. The authors would like to acknowledge the support obtained by the team from Airbus Defence \& Space, specially we would like to acknolwedge the  Savier Open Innovation project members: Gemma Blasco,  C\'esar Castro, and Jos\'e Insenser.

\ifCLASSOPTIONcaptionsoff
  \newpage
\fi



\bibliographystyle{IEEEtran}
\bibliography{IEEEabrv,mybib}
\end{document}